\pdfoutput=1

\documentclass[11pt]{article}

\usepackage{emnlp2022}

\usepackage{times}
\usepackage{latexsym}

\usepackage[T1]{fontenc}

\usepackage[utf8]{inputenc}

\usepackage{microtype}

\usepackage{colortbl}
\usepackage{graphics}
\usepackage{placeins}
\usepackage{graphicx}
\usepackage[caption=false]{subfig}

\usepackage{xspace}
\usepackage{enumitem}

\newcommand{\todo}[1]{}

\usepackage[htt]{hyphenat}
\usepackage{xcolor}
\newcommand{\red}[1]{\textcolor{red}{#1}}

\newcommand{\code}[1]{{\small{\textmd{#1}}}}
\newcommand{\tinycode}[1]{{\tiny{\textmd{#1}}}}

\usepackage{amsfonts}

\newcommand{\Facts}{\ensuremath{\mathcal{F}}\xspace}
\newcommand{\Props}{\ensuremath{\mathcal{P}}\xspace}
\newcommand{\Ops}{\ensuremath{\mathcal{O}}\xspace}

\newcommand{\pmOset}[2]{\mathrm{#1}(#2)}
\newcommand{\Opre}[1]{\ensuremath{\pmOset{pre}{#1}}\xspace}
\newcommand{\Oeff}[1]{\ensuremath{\pmOset{eff}{#1}}\xspace}
\newcommand{\Oneff}[1]{\ensuremath{\pmOset{neff}{#1}}\xspace}

\newcommand\quasi{\char`\~}

\usepackage{listings}
\definecolor{codegreen}{rgb}{0,0.6,0}
\definecolor{codegray}{rgb}{0.5,0.5,0.5}
\definecolor{codepurple}{rgb}{0.58,0,0.82}
\definecolor{backcolour}{rgb}{0.95,0.95,0.92}

\lstdefinestyle{quote_style}{
    backgroundcolor=\color{backcolour},   
    commentstyle=\color{codegreen},
    keywordstyle=\color{magenta},
    numberstyle=\tiny\color{codegray},
    stringstyle=\color{codepurple},
    basicstyle=\rmfamily\small,
    breakatwhitespace=false,     
    columns=flexible,
    breaklines=true,                 
    captionpos=b,                    
    keepspaces=false,                 
    numbers=left,                    
    numbersep=2pt,                  
    showspaces=false,                
    showstringspaces=false,
    showtabs=false,                  
    tabsize=1
}

\lstdefinestyle{kb_style}{
    backgroundcolor=\color{backcolour},   
    commentstyle=\color{codegreen},
    keywordstyle=\color{magenta},
    numberstyle=\tiny\color{codegray},
    stringstyle=\color{codepurple},
    basicstyle=\rmfamily\small,
    breakatwhitespace=false,     
    columns=fullflexible,
    breaklines=true,                 
    captionpos=b,                    
    keepspaces=false,                 
    numbers=left,                    
    numbersep=2pt,                  
    showspaces=false,                
    showstringspaces=false,
    showtabs=false,                  
    tabsize=1
}

\title{Improving Generalization in Task-oriented Dialogues \\ with Workflows and Action Plans}

\author{
 	Stefania Raimondo \\
 	\small{\textmd{stefania.raimondo@servicenow.com}} \\
 	\And
 	Christopher Pal \\
 	\small{\textmd{christopher.pal@servicenow.com}} \\
 	\And
 	Xiaotian Liu\thanks{\hspace{0.5em}Work done when Xiaotian Liu was an intern at ServiceNow Research.} \\
	\small{\textmd{xiaotian.liu@mail.utoronto.ca}} \\
 	\AND
 	David Vazquez \\
 	\small{\textmd{david.vazquez@servicenow.com}} \\
 	\And
	Hector Palacios \\
	\small{\textmd{hector.palacios@servicenow.com}} \\
}

\begin{document}
\maketitle
\begin{abstract}
Task-oriented dialogue is difficult in part because it involves understanding user intent, collecting information from the user, executing API calls, and generating helpful and fluent responses. However, for complex tasks one must also correctly do all of these things over multiple steps, and in a specific order.
While large pre-trained language models can be fine-tuned end-to-end to create multi-step task oriented dialogue agents that generate fluent text, our experiments confirm that this approach alone cannot reliably perform new multi-step tasks that are unseen during training.
To address these limitations, we augment the dialogue contexts given to \textmd{text2text} transformers with known \textit{valid workflow names} and \textit{action plans}. 
Action plans consist of sequences of actions required to accomplish a task, and are encoded as simple sequences of keywords (e.g. verify-identity, pull-up-account, reset-password, etc.). 
We perform extensive experiments on the Action-Based Conversations Dataset (ABCD) with T5-small, base and large models, and show that such models: a) are able to more readily generalize to unseen workflows by following the provided plan, and b) are able to generalize to executing unseen actions if they are provided in the plan. In contrast, models are unable to fully accomplish new multi-step tasks when they are not provided action plan information, even when given new valid workflow names.   
\end{abstract}

\section{Introduction}
\begin{figure}
    \includegraphics[width=0.5\textwidth]{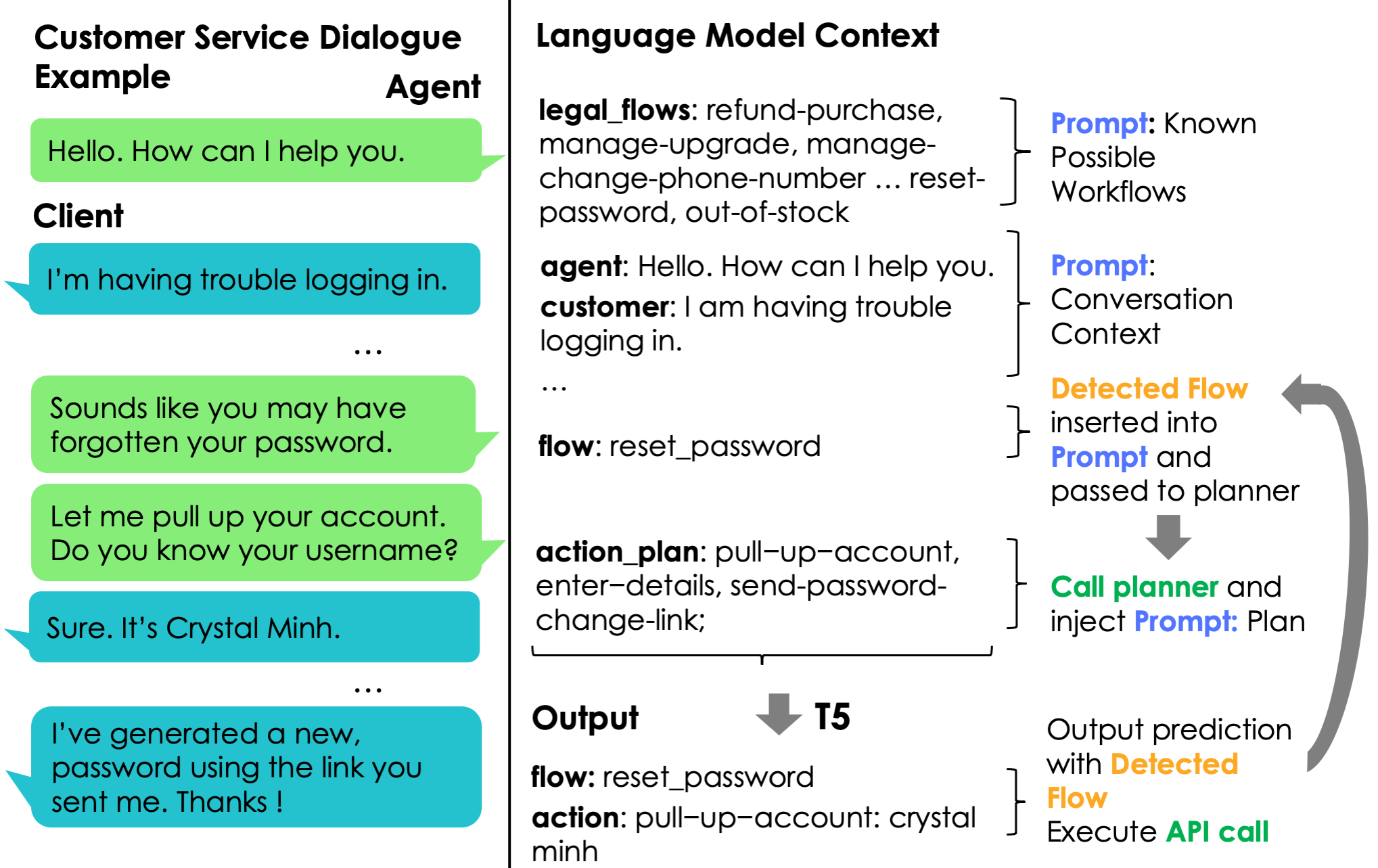}
    \caption{\small Our methodology is to drive a large language model (LLM) through prompts and interactions with an external planning mechanism. Initially, the prompt may consist of a list of valid workflows. When the user intent is detected, the planner feeds the LLM with a plan that helps guide the task-oriented dialog. This approach permits new workflow steps to be added on-the-fly, allows for one to have dynamically restructured workflows, and to update plans to reflect only un-executed steps as actions are taken by the agent. In our experiments, we preprocess the data to simulate this interaction, allowing us to evaluate how plan-based prompting can enhance generalization to new multi-step tasks not seen during training.
}
    \label{fig:idea_overview}
\end{figure}

Dialogue allows service agents and users to accomplish complex tasks flexibly and naturally. However, such dialogues are challenging for automated agents because success requires the tracking of long-range information and correct behaviour at every step of multi-step tasks.
In service-focused task-oriented dialogue, an agent and a user interact back-and-forth with natural language text to reach a goal determined by the user. The agent must identify the task the user intends to solve, collect relevant information from the user, and execute actions until the task is complete. The possible set of actions and the order in which they  are accomplished depends on the specific task and environment.

Recent work has applied modern large language models (LLMs), e.g. \citet{raffel_exploring_2020,brown_language_2020}, etc., to complex structured reasoning tasks including task-oriented dialogue%
~\cite{hosseini-asl_simple_2020, peng_soloist_2021, he_galaxy_2022, peng_soloist_2021, ham_end--end_2020, gao_robust_2020}. 
By leveraging large-scale pretraining,
these approaches can generate more fluent text and better interpret natural language. However, our experiments, along with recent work~\cite{valmeekam_large_2022, bender_dangers_2021, thoppilan_lamda_2022}, find that pure language model solutions cannot reliably solve complex collaborative tasks, especially when presented with novel goals or types of information, or when the steps required must satisfy external constraints like business rules.

We hypothesize that performance on task-oriented dialogue can be improved by augmenting language models with prompts. These prompts provide explicit information outlining possible workflows and action plans for accomplishing a given dialogue task. At the same time, the model must engage in an interleaved dialogue with the user and handle actions. We finetune a T5 text2text model~\cite{raffel_exploring_2020} on the multi-step tasks of the Action Based Conversations Dataset (ABCD) \cite{chen2021action}, augmenting the dialogue context provided to the model with explicit plans describing the required remaining action sequence necessary to correctly execute the required workflow and solve the user's problem. Such a text2text setup is flexible
as
new information can be provided to the model at test time to describe the required outputs, as opposed to relying on a traditional classifier for every new task. We then evaluate the ability of different model variants to execute new steps and tasks unseen during training for different levels of difficulty or novelty with respect to the training distribution. %

Our contributions are as follows:
\begin{enumerate}[noitemsep,topsep=0pt]
    \item We show that LLMs struggle to generalize to unseen action plans. 
    \item Our proposed text2text models %
    can generalize better to new workflows and new tasks, executing actions not seen during training %
    by having 
    access to plans for these new tasks.
    \item We evaluate our approach on the ABCD dataset and propose new metrics that better fit the evaluation of plan execution.
\end{enumerate}

The most dramatic impact of adding action-plans to the prompt is for actions that \textit{do not} appear in the training data, a extreme challenge for generalization.
Table~\ref{tab:large-result-split-3-seen-vs-unseen}, in section~\ref{sec:unseen-actions},
shows that adding the plan raises action accuracy from 0\% to \quasi44-69\%.

\section{Task-oriented Dialogues \& ABCD}

In our experiments we use the Action Based Conversations Dataset (ABCD) of \citet{chen2021action}, which consists of 10k multi-step task based dialogues between two humans with 55 different high level intents. Each task requires specific sequences of actions to be completed to successfully achieve a goal. Dialogues in ABCD have an online retail theme and involve tasks like resetting passwords and checking the status of orders. One person %
plays the role of an agent, the other plays the role of the client. Agents interact with a simulated computer system that requires different types information from the user to execute different types of actions such as resetting passwords, and checking orders. Agents are provided with and told to follow pre-defined \textit{workflows} consisting of sequences of steps or actions that must be taken to perform a given task. Actions may have slots and optional slots (i.e. parameters), the values of which are gathered from %
from the user. 
Having multi-step tasks composed of multiple actions, each with their own slots makes ABCD unique as a task-oriented dialogue dataset. 

We obtain the desired sequences of actions for each workflow from the database provided by the ABCD dataset\footnote{\href{https://github.com/asappresearch/abcd/blob/master/data/kb.json}{github.com/asappresearch/abcd/blob/master/data/kb.json}}.
However, customer agents deviate from such sequences for good reasons like some actions being optional depending on the context, or while uncovering the user's intent and the correct workflow.

\citet{chen2021action} both introduced the ABCD dataset and performed baseline experiments using models based on BERT \cite{devlin_bert_2018}, Al-BERT \cite{lan2019albert}, RoBERTa \cite{liu2019roberta} and RoBERTa-Large.
Unlike their work, we use a 
T5
text2text model %
which we influence by conditioning on worfklow plans,  %
as a form of prompt or plan conditioning. One of our key goals is to understand if these these plans are able to 
improve generalization to \emph{new} workflows. As such, we use three new alternative splits of the data in which different workflows, action sequences, and actions are not seen during training.

\subsection{Actions, Slots, Goals and Workflows}
\label{sec:task-oriented-dialogue}

Formally, we consider a dialogue $D$ composed of a total of $T$ ordered turns $t_i$: $D = t_1, \ldots t_T$. %
A turn, $t_t = (s_t, u_t, a_t)$, has a speaker, either the agent $s_t=S_A$ or the user $s_t=S_U$,
an utterance $u_t$ which is a natural language string, and an action $a_t$. For ABCD, one of either the action or utterance is null.

We assume the agent must identify the goal or intent of the user, $i_t$, and execute an ordered set of actions, $a^1, \ldots, a^k$, with $k<T$, to accomplish this goal.
The actions can use \textit{slots (arguments)} consisting of information extracted during the dialogue, available in a database, or produced by another action.
A goal is %
achieved if its complete set of actions is executed in the correct order with correct arguments. If the ordered set of actions follow a fixed process, it is called a \textit{workflow}. 
We use the terms \textit{workflow} and \textit{flow} to refer to the goal or goal name, \textit{action sequence} for the corresponding sequence of actions, especially if it is shared across workflows, and \emph{action plan} for the sequence of remaining actions for the workflow. %
In appendix~\ref{sec:planning-dialogues} we comment on how symbolic planning \cite{geffner2013concise} can be used to generate these workflows and the prompts used in this paper. However, in our experiments here action plans are stored as a lookup-table.

\section{Related Work}

We make use of the text2text paradigm,  finetuning and augmenting context provided to T5 models.
We find that this provides good performance in practice while remaining small enough to allow for more extensive experiments. 
Recent work has also demonstrated that the zero shot performance of large language models trained to perform multiple tasks can be improved by providing instructions \cite{wei2021finetuned}. 
However, their work does not examine instructions for complex multi-step tasks and/or dialogues and language models alone cannot reliably tackle such tasks~\cite{valmeekam_large_2022,bender_dangers_2021,thoppilan_lamda_2022}.

The traditional approach to building task-oriented dialogue systems is to build a pipeline of modules: %
an NLU component for identifying user intents and entities, a dialogue manager for keeping track of state, executing and planning actions, and an NLG component for generating output utterances. Most commercial systems, like Rasa\footnote{\href{https://rasa.com}{https://rasa.com}}, Watson\footnote{\href{https://cloud.ibm.com/catalog/services/watson-assistant}{https://cloud.ibm.com/catalog/services/watson-assistant}}, Dialogflow\footnote{\href{https://cloud.google.com/dialogflow}{https://cloud.google.com/dialogflow}}, and Power Virtual Agent\footnote{\href{https://powervirtualagents.microsoft.com/en-ca/}{https://powervirtualagents.microsoft.com/en-ca/}} rely on such a pipeline. Building such systems is typically costly, involving handcrafting of templates, intents, and dialogue trees and providing specific examples of each step in the dialogue. Work on these systems typically focuses on how to make their development easier with smarter dialogue managers~\cite{xie_converse_2022, machines_task-oriented_2020}. 
Recently, \citet{muise2019planning} propose an alternative to fully-specified dialogue trees based on the use of symbolic planning to guide both low-level and domain actions. APIs, their requirements and possible results are interpreted as planning actions. Using a planning full-observable non-deterministic problems, they synthesize the possible behaviour of the dialogue. However, their approach does not involve the use of modern large neural language models.
We would like to strike a balance between the rigidity of hand crafted dialogue managers and the flexibility of neural approaches.

 An alternative approach to dialogue modelling is to train a unified system end-to-end~\cite{wen_network-based_2017, vinyals_neural_2015, bordes_learning_2017} with the aim of removing the burden of hand-crafting dialogue trees and eliminating error-propagation in a waterfall-style system. The main issue with these methods is a lack of robustness: task-oriented dialogues must execute well-structured database calls, correctly execute tasks, and be generally informative and helpful. However, these models are trained to parrot existing dialogue responses, with no concept of higher-order conversational goals~\cite{vinyals_neural_2015}. Furthermore, early work in this direction was limited by the underlying models (typically LSTMs~\cite{wen_network-based_2017} or GRUs~\cite{shang_neural_2015, shu_incorporating_2018} and MemoryNetworks~\cite{bordes_learning_2017}) which struggled to produce diverse and relevant utterances~\cite{li_diversity-promoting_2015} and keep track of conversation history. 

Recent work has proposed a Description-Driven Dialog State Tracking (D3ST) approach to task oriented dialogues \cite{zhao2022description}. In their framework the input to an encoder is augmented to include slot descriptions, intent descriptions, and conversation context concatenated into a single string. In contrast to our work, they do not inject full plans for complex multi-step workflows.

Other work has focused on modelling belief-state and/or slot-value information, 
for example work such as ~\citet{hosseini-asl_simple_2020, peng_soloist_2021, he_galaxy_2022, peng_soloist_2021, ham_end--end_2020}, and \citet{gao_robust_2020} trained GPT/GPT-2 or UniLM style models to output belief state (i.e. the state of known slot-filling information), next-action, and the output utterance utterance, by providing the model with dialogue context and (optionally) the summarized Database output. ~\citet{peng_soloist_2021} also pretrain across multiple task-oriented dialogue datasets, showing improved performance and use iterative human-in-the-loop training on the final domain. ~\citet{he_galaxy_2022} instead augment pretraining with a semi-supervised consistency objective. However, importantly these works all focus on the MultiWoz~\cite{budzianowski_multiwoz_2020} dataset, 
which does not contain ABCD's features like multi-step constraints that must be respected to complete tasks.

High-quality task-oriented dialogue datasets are difficult to collect and so there are relatively few in the literature. Early examples include ATIS~\cite{hemphill_atis_1990} and DSTC2~\cite{henderson_second_2014}, while some of the most widely used more modern datasets include Simulated Dialogue (M2M)~\cite{shah_building_2018}, MultiWoz~\cite{budzianowski_multiwoz_2020}, TaskMaster~\cite{byrne_taskmaster-1_2019}, Frames~\cite{asri_frames_2017}, and the Schema-Guided Dialogue Dataset~\cite{rastogi_towards_2020}. MultiWoz is perhaps the most popular. However, neither SGD nor Multiwoz require actions to be executed in a particular order; intents map directly to actions and the agent's task is only to request the necessary slots, instead of executing a \textit{sequence} of actions.
We focus our experiments on the ABCD dataset and problem formulation because, unlike these other datasets ABCD involves task oriented dialogues that require following a multi-step workflow.  
See Appendix~\ref{sect:related-extended} for further discussion of related work.

\section{Experimental Setup}

\subsection{Dataset}
We use the ABCD dataset for both training and testing our models. ABCD dialogues centre around the completion of workflows (i.e. goals or intents) which are associated with sequences of actions. Note that some workflows have the same action sequence but achieve different goals (shown in Table \ref{tab:kb_split_all} in Appendix~\ref{appendix:dataset_splits}). Instructions for completing workflows (i.e. business logic) are provided to the human agent during dataset collection as phrases in natural language.  Importantly, these instructions include "if-else" type logic, requiring the agent to adapt to the information provided by the user. However, this branching is not explicit in the knowledge base we use for our experiments, which describes workflows as fixed sequences of actions. 
See Appendix~\ref{appendix:abcd_dataset} for an example of the instructions provided to the human agent.

ABCD contains 55 workflows (intents or goals), 30 unique actions (essentially API-calls) with 231 unique slots, 12k unique slot values, and \quasi22 turns per conversation.  For comparison, Multiwoz~\cite{budzianowski_multiwoz_2020} contains 7 actions\footnote{called \textit{domains} in Multiwoz}, 24 unique slots, 4.5k unique slot values, and \quasi13 turns per conversation. Slot values can be of two varieties: enumerable values (e.g. \code{membership\_level} can be "guest", "bronze", etc.) or non-enumerable (e.g. street\_address is a freeform string). 
During data collection, the
slot values come either from the enumerable values known to the agent or the dialogue context. ABCD provides both the original and delexicalized versions of the dialogues.  %
Appendix~\ref{sec:dataset-issues} comments on issues of the ABCD dataset.

Importantly, actions in ABCD have optional arguments. %
For example, \code{verify-identity} requires any 3/4 slots, \code{validate-purchase} requires all 3/3 slots, and the rest require only one of their possible slots.

\subsection{Text2Text Task Formulation}

We perform task-oriented dialogue by re-framing the task as seq2seq (or text2text) prediction with a text-generation model. We use an encoder-decoder T5 model~\cite{raffel_exploring_2020}, but any model which predicts an output string $o_t$ given an input context string $c_t$ would do. The context string contains the dialogue history, which is a concatenation of the string representations of all previous turns (utterances or actions) in the dialogue: $c_t = t_1, \ldots, t_T$. We use the following format for each turn, where turn type is either "action", "agent" or "customer": \code{<turn\_type>: <utterance or action> }. Actions are inserted into the context in the following format: \code{<action>:  <slot\_value\_1>, <slot\_value\_2>...;} which provides only the value of the slots, and not the slot names (e.g. phone number) or slot types (e.g. 10-digit number). An example conversation (and context) is provided in Fig \ref{fig:idea_overview}. 
The basic output predicted by the model is of the same format as the input. Note also that these contexts are drawn from the original (non-delexicalized) data. %

\subsection{Augmenting the context}
\label{sec:augmenting_context}

As discussed above, we augment the prompting context of our model with information useful for guiding the model. We experiment with three types of prompting formatted as follows:
\begin{enumerate}[noitemsep,topsep=0pt] %
    \item \textbf{workflow names (F)}: \code{flow: <workflow\_name>;}
    \item \textbf{action-plans (P)}, i.e. the list of \textit{remaining} actions required to complete the workflow:  \code{action\_plan: <action\_1>, <action\_2>, ...;} 
    \item \textbf{lists of legal (known) workflow names (L)}:
 \code{legal\_flows: <workflow\_name>, ...;}
\end{enumerate}

Note that action plans are always preceded by the associated workflow label. The codes in the brackets above (i.e. F, P, L) are used to specify the model variants that we have explored in Table \ref{tab:large-result-overall}.

The legal flows (L) are provided at the beginning of the dialogue context, the flow name (F) and action plan (P) at the end.
We also experimented with adding the extra information before every turn in the context and before/after action turns, but exclude these results.

As we discuss in Section \ref{sec:generalization_splits}, we created new splits of the ABCD data such that different types of workflows are removed from the training data and only exist at test time. 
As such %
the list of known workflow names (L) differs during testing and training -- only those present in the training set are provided at training, while the full list is provided at test time.

Along with the prediction of the action or utterance, we train the model to predict the workflow name, in the same format described above. %

\subsection{Implementation Details}

During training, we use teacher-forcing, meaning that we treat each turn in a conversation as an independent training example, providing the correct context to the model (augmented as necessary) at each turn and ignoring the model's predictions for previous turns. Thus, errors made by the model are not propagated to future turns. This is important for training since a) customer utterances are static and b) it allows efficient batching of examples.

For every setup we train a new model, though in practice there may be benefits from multi-stage training or transfer learning.

We use the T5 Pytorch implementation provided by huggingface (hf)\footnote{\href{https://huggingface.co/}{https://huggingface.co/}} and train using the hf library. We use the hf pretrained T5-small, T5-base, and T5-large models (T5ForConditionalGeneration) with 60, 220, and 770 million parameters, respectively. We use the default hyperparameter settings from hf, including: AdamW optimizer ($\beta_1=0.9, \beta_2=0.999, \epsilon=1e-8$) without weight decay, learning rate of $5e-5$, linear scheduler without warmup, max grad norm of $1.0$, beam search of 5, and we train for 3 epochs. 

Given that plans do not change throughout a conversation in ABCD (although the remaining steps do decrease), a simple lookup table can be used as a planner. However, 
we have also implemented the planning model in the \textit{Planning Domain Definition Language}  (PDDL)\footnote{\href{https://planning.wiki/}{https://planning.wiki/}}\cite{mcdermott-et-al-tr1998} which allows for new workflows to be easily added, plans with branches to be included, and plans to be dynamically updated throughout a conversation.
In Appendix~\ref{sec:planning-detail} we show how adding new actions and constraints could affect many workflows.

\section{Evaluation Methodology}
\label{sec:eval_method}
\label{sec:generalization_splits}

To evaluate generalization ability, we create three new train/test splits of the ABCD dataset, in which some flows are seen only at test time. 

In the ABCD ontology, workflows are divided into semantically related workflow groups\footnote{In ABCD, workflows and workflow groups are actually referred to as "subflows" and "flows", respectively. We change our terminology to align with the concept of a workflow.} and the names of workflows within a group may share a prefix. For example, the \code{manage\_account} group contains \code{status\_service\_added}, \code{status\_service\_removed}, \code{manage\_change\_name} %
among other flows. Flows which share a prefix ("workflow prefix groups") (e.g. \code{status\_} %
in the previous example), are more likely to share or partially share action sequences. %

We use the following strategies to create our new dataset splits.
\begin{itemize}[noitemsep,topsep=0pt]
\item 
\textbf{Standard:} All flows and actions are seen during training and testing. %
\item 
\textbf{Split~1:} No flows in testing are seen during training. Approximately half of each workflow prefix group is split into train/test.
\item 
\textbf{Split 2:} No action sequences in testing are seen in training. Approximately half of each workflow prefix group is split into train/test.
\item 
\textbf{Split 3:} No action sequences or workflow prefix groups in testing are seen in training. %
\end{itemize}
Note that these splits result in very different test sets, in terms of size and contained flows. %
The division of workflows into train and test and the number of examples in each are provided in Tables ~\ref{tab:kb_split_all} and ~\ref{table/n_turns_per_split}. %
Note that this indicates the \textit{theoretical} distribution of actions. Agents do not always follow the guidelines and execute the labelled workflow steps. %
For example, the following eight actions are only in the testset workflows in Split 3, but appear at least once in training: \code{instructions}, \code{make-password}, \code{search-jeans}, \code{search-shirt}, \code{search-jacket}, \code{search-boots}, \code{log-out-in}, \code{try-again}, \code{subscription-status}.

Since we use a text2text paradigm, our models' outputs are strings. Thus, we parse the outputs to extract predicted actions, slots etc. For every turn, we run a regex for the expected output format for the expected turn type (i.e. utterance or action)\footnote{For example, for action turns for model +F+P, we use the following regex:  {\tinycode{(?P<flow\_label>flow:)(?P<flow>.*);\textbackslash s*} \tinycode{(?P<action\_label>action:)(?P<action>.*):(?P<slots>.*)}}. For utterance turns, we use \tinycode{(?P<flow\_label>flow:)(?P<flow>.*);\textbackslash s* (?P<agent\_text\_label>agent:)(?P<agent\_text>.*)}}. 
Thus, incorrectly formatted outputs are deemed incorrect. %

We consider the following metrics:
\textbf{Action}: action accuracy, average of per-turn action exact match (string match) for ground-truth action-turns only. %
    \textbf{Flow}: flow accuracy, average of per-turn flow exact match.
    \textbf{Flow prefix}: flow prefix accuracy, average of per-turn flow-prefix (e.g. \code{status\_} for flow \code{status\_service\_added}) exact match.
    \textbf{Lev act/free del}: Levenshtein distance required to transform the predicted action sequence to the ground-truth sequence. All predicted actions are extracted from all turns in a dialogue, including non-action turns. Addition, substitution, and deletion have a cost of 1. For the "free del" version, deletion has cost 0, and %
    might be more tolerant to justified deviations from the prescribed action sequence. %
We compute our metrics using the average of testing three models trained with separate seeds.
Though models are trained to output slot values, we relegate slot accuracy metrics to Appendix~\ref{sec:slot_metrics} to focus on action and flow prediction. For current models, %
slot accuracy %
does not differ significantly across dataset splits.

The definition and computation of our metrics above are slightly different from the original ABCD evaluation which contains some noise and ambiguity. 
However, a T5 based model for the ABCD has been evaluated in \citet{hattami2022workflow} using exactly the same metrics as the original ABCD evaluation and it yielded the results (reproduced from their work) in Table 
\ref{tab:ast}, showing state of the art (SOTA) or comparable performance to the previous result of \citet{chen2021action}. Thus, given that T5-small produces SOTA results on the original ABCD evaluation, we perform our experiments with T5-based models, and experiment with larger model variants (which consistently increase performance on this benchmark across different metrics).

\begin{table}[b]
\begin{center}
\resizebox{0.48\textwidth}{!}{
\begin{tabular}{p{4cm}ccl}
\hline

 \textbf{Model}                        & \textbf{B-Slot} & \textbf{Value}  & \textbf{Action} \\\hline
ABCD-RoBERTa     & \textbf{93.6\%} & 67.2\%          & 65.8\%  \\
ABCD-T5-Small       & 89.1\%          & \textbf{89.2\%} & \textbf{87.9}\%\\
\hline
\end{tabular}
}
\end{center}
\caption{\label{tab:ast} ABCD-T5 results on the ABCD test dataset using the original evaluation metrics (reproduced from \citet{hattami2022workflow}). %
In \cite{chen2021action}, %
B-Slot and Value metrics measure accuracy for action \& value prediction, respectively. The Action metric is joint accuracy (i.e. correct prediction of both %
B-Slot \& Value.)}
\end{table}

Our setup and metrics are designed to be more sensitive and realistic than \citet{chen2021action} and \citet{hattami2022workflow}, through addressing two issues. First, we use the non-delexicalized data, which is free from delexicalization errors. Second, we predict both actions and utterances, meaning the agent must also predict whether to perform an action or utterance on a given turn, which accounts for a large portion of action prediction errors. Executing the right action at the right moment is a fundamental problem for customer agents. In comparison to the original ABCD setup and evaluation our setup, involving predicting both actions and utterances and the corresponding changes in metrics, is simply harder. Correspondingly, models score slightly lower on our Action metric compared to the original ABCD metric.

\section{Results}
In our experiments, we aim to show that explicitly adding a plan of remaining action sequences to the dialogue context as well as lists of allowable workflow names and the current goal/workflow of the conversation improves zero-shot action and flow prediction performance.

We provided results for our T5-large based models in Table \ref{tab:large-result-overall}. Results for T5-small and T5-base are provided in Appendix \ref{t5_small_results}.
In general, T5-small metrics are \quasi5-10\% lower than for T5-base, which are \quasi3-7\% lower than for T5-large. As one can observe from Table \ref{tab:ast}, even our baseline finetuned T5-small model variants serve as a strong starting point for subsequent experiments. However, we begin our analysis by first observing that the finetuned T5 baselines completely fail to correctly follow workflows for Splits 1-3, with 0\% on our Flow correctness metric. %

\begin{table}[h!]
    \small
    \centering
    \resizebox{0.5\textwidth}{!}{
    \centering
    \begin{tabular}{p{2.9cm} p{0.6cm} p{0.7cm} p{0.8cm} p{1.1cm}}
        \hline
        Model + Additional Context & Action & Flow & Flow prefix & Lev act / free del \\
        \hline
         \multicolumn{5}{l}{\textbf{Standard split}: all actions and workflows seen in training} \\
        \hline
        T5-large finetuned & 85.1 & 85.8 & 90.1 & 2.8 / 0.3 \\
        + L & 85.2 & 85.7 & 89.9 & 2.7 / 0.3\\
        \arrayrulecolor{lightgray}\hline
        + F & 85.2 & 93.5 & 94.5 & 2.7 / 0.3\\
        + L + F & 85.4 & 93.5 & 94.5 & 2.8 / 0.3\\
        \arrayrulecolor{lightgray}\hline    
        + F + P & 85.3 & 93.5 & 94.5 & 2.9 / 0.3\\
        + L + F + P & 85.4 & 93.5 & 94.5 & 2.8 / 0.3\\
        \arrayrulecolor{black}\hline
        \multicolumn{5}{l}{\textbf{Split 1}: All workflows in results below are unseen in training} \\ 
        \arrayrulecolor{black}\hline
        T5-large finetuned  & 77.8 & 0.0 & 41.5 & 2.9 / 0.6\\
        + L & 77.7 & 27.2 & 56.8 & 2.9 / 0.6\\
        \arrayrulecolor{lightgray}\hline 
        + F & 77.4 & 93.0 & 94.6 & 2.9 / 0.6\\
        + L + F & 77.5 & 93.4 & 94.6 & 2.9 / 0.6\\
        \arrayrulecolor{lightgray}\hline 
        + F + P & 83.7 & 93.4 & 94.6 & 2.9 / 0.4\\
        + L + F + P & 83.0 & 93.4 & 94.6 & 2.9 / 0.4\\
        \arrayrulecolor{lightgray}\hline 
        
        \arrayrulecolor{black}\hline
         \multicolumn{5}{l}{\textbf{Split 2}: Flow prefix groups \& action seq. unseen in training} \\
        \arrayrulecolor{black}\hline
        T5-large finetuned & 75.2 & 0.0 & 24.7 & 3.5 / 0.7\\
        + L  & 75.7 & 20.5 & 41.7 & 3.5 / 0.7\\
        \arrayrulecolor{lightgray}\hline 
        + F & 74.0 & 92.0 & 94.4 & 3.3 / 0.8\\
        + L + F & 75.2 & 91.5 & 94.3 & 3.4 / 0.7\\
        \arrayrulecolor{lightgray}\hline 
        + F + P  & 81.5 & 92.5 & 94.4 & 3.3 / 0.5\\
        + L + F + P & 81.5 & 93.7 & 94.4 & 3.3 / 0.5\\
        \arrayrulecolor{black}\hline
        
         \multicolumn{5}{l}{\textbf{Split 3}: Flow prefix groups \& action seq. unseen in training} \\
        \arrayrulecolor{black}\hline
        T5-large finetuned & 60.0 & 0.0 & 16.9 & 3.7 / 1.1\\
        + L & 59.7 & 27.1 & 46.4 & 3.3 / 1.1\\
        \arrayrulecolor{lightgray}\hline
        + F & 60.9 & 92.7 & 94.0 & 3.5 / 1.0\\
        + L + F & 61.6 & 93.2 & 93.7 & 3.4 / 1.0\\
        \arrayrulecolor{lightgray}\hline 
        + F + P & 69.6 & 92.9 & 93.9 & 3.0 / 0.7\\
        + L + F + P & 70.2 & 93.2 & 93.5 & 3.2 / 0.7\\ 
        \arrayrulecolor{black}\hline
    \end{tabular}
    }
    \caption{Action, slot, and flow prediction metrics for models trained across various generalization splits explained in
Section \ref{sec:eval_method}. The Baselines are finetuned T5-large models, and model variants provide different amounts of additional context beyond
the dialogue utterances described in Section \ref{sec:augmenting_context}, consisting of: legal flow lists (L), flow (last turn) (F), or the action sequence plan
(last turn) (P)}
    \label{tab:large-result-overall}
\end{table}

\subsection{Action prediction}

On Splits 1-3 we observe that
once unseen flows appear at test time, action accuracies differ substatially across the different context configurations. On Split 1 when test flows are unseen in training, without flow information action prediction drops to 77.8\%. %
Providing the flow name alone does not improve action accuracy (77.4\%) %
(i.e. +F), but also providing the plan (+F+P) does improve action accuracy to 83.7\%. %
Further providing the legal list of flows (i.e. +L) does not generally provide a boost for T5-large in Split 1.

For Split 2, when all action sequences in test are unseen during training, action prediction accuracy drops by another \quasi3\%. %
The gap between models without the plan and with the plan remains at \quasi5-6\%. %
Minimal benefit for this split exists by providing the legal list of flows. %

For Split 3, when all action sequences in the test are unseen in training and entire flow-prefix groups are also unseen, action accuracy is much higher for models given plans (i.e. +F+P, +L+F+P) at \quasi70\% %
as compared to models with only the flow (i.e. +F, +L+F) at \quasi61-61.5\%, %
and with neither flow nor plan (i.e.+$\emptyset$, +L) at \quasi60\%, %
a difference of \quasi10\%. The benefit to providing the legal list of flows is very minor, with only <1\% improvement for the model that is given a plan (i.e. +F+P vs +L+F+P). Similarly, models given the plan have lower Levenshtein action distances. %

\subsection{Unseen actions}
\label{sec:unseen-actions}

\begin{table}[h!]
    \centering
    \resizebox{0.5\textwidth}{!}{
    \centering
    \begin{tabular}{l p{3cm} p{2.5cm}}
        \hline
        Model & Theoretical seen / unseen actions & Actual seen /  unseen actions \\
        \hline
        T5-large finetuned & 65.0 / 37.4 & 61.2 / 0.0   \\
        + L  & 64.6 / 37.5 & 60.8 / 0.0 \\
        + F & 66.1 / 37.2 & 62.1 / 0.0  \\
        + L + F & 66.6 / 39.0 & 62.9 / 0.0  \\
        + F + P & 74.0 / 50.0 & 70.1 / 44.4\\
        + L + F +P & 73.7 / 54.8 & 70.3 / 69.8 \\
        \hline
        \end{tabular}
        }
    \caption{Action accuracy for theoretically/actually seen/unseen actions in training on T5-large for Split 3}
    \label{tab:large-result-split-3-seen-vs-unseen}
\end{table}

Table \ref{tab:large-result-split-3-seen-vs-unseen} shows the action prediction for actions which were seen vs unseen in the training set for Split 3, with a column showing the theoretically vs actually seen/unseen actions (see Section \ref{sec:generalization_splits}).

As shown in Table \ref{tab:large-result-split-3-seen-vs-unseen}, as expected, models that are not provided the plan are unable to generalize to unseen actions in Split 3. Models which are given the plan, however, (i.e. +F+P, +L+F+P) have action accuracies of \quasi45\% and \quasi70\% %
for \textit{actually unseen} actions. Models without plans predict the never before seen actions with \quasi0\% accuracy. This suggests that the models are \textit{using} the actions from the plans, rather than trying to infer the actions exclusively based on the dialogue context. 

Even on actions actually seen in the training set, models with plans (+F+P, +L+F+P) have a \quasi8-10\% %
increase in action accuracy. A similar improvement exists for actions that should have theoretically been seen/unseen in training of \quasi8-10\%/\quasi10-12\%. %

Interestingly, for models given the plan, action accuracy is actually higher for actually unseen actions than for actually seen actions.  This might reflect a difference in difficulty between actions.

\subsection{Specific action errors}
We provide confusion matrices for +F+P T5-large models trained on Split 3 with different seeds in Figures~\ref{fig:action_conf_matrix_large_d31-1_seed0_split3}, \ref{fig:action_conf_matrix_large_d31-1_seed1_split3}, and \ref{fig:action_conf_matrix_large_d31-1_seed2_split3} of Appendix~\ref{sec:action-prediction-confusion-matrices}. The errors made by these models are not identical. 

The only action not actually present in training is \code{make-password} which has varied performance across all models (from 30\% to 60\%) but is consistently confused with \code{enter-details} (the action preceding it in all flows) or \code{verify-identity}. %

Other actions that are not present in plans in training have wildly different accuracies. For example, \code{instructions} has \quasi0\% accuracy across all models, and is confused with \code{try-again}, \code{log-in-out} (both co-occurring in flows), and \code{<blank>}, while \code{search-jacket} has 90 or 100\% accuracy and no consistent errors. %
\code{<blank>} is one of the most common errors for many actions. This occurs when the model predicts an utterance turn instead of an action turn. 

There are also a few examples of incorrectly formatted actions or totally new actions (e.g. \code{offer-discount}, \code{call-centre}, \code{notify-team: placing-order}, etc.)%
, which suggest that the model \textit{is} capable of inventing new actions, perhaps for  unusual conversations.

The above suggests that incorrect actions are often the result of predicting an action on the wrong turn (i.e. \code{<blank>}) or following the flow's actions when they do not follow the ground-truth labels. 
See Appendix~\ref{sec:example-unseen-action-correct} for an example where
an unseen action is correctly executed by the agent, the name of which is drawn from the provided plan. %
Appendix~\ref{sec:example-woz-does-not-follow-flow} shows another case where the agent follows the plan for an unseen flow and action sequence. In this example, however, action-accuracy is very low because the \textit{ground-truth} actions \textit{do not follow} the ground-truth flow. Further investigation is required to determine the extent of this ground-truth action/flow mismatch.

\subsection{Flow and Flow prefix prediction}
\label{sec:flow-prefix-prediction}
All models which receive flow information (all except +$\emptyset$ and +L) achieve >\quasi91\% flow prediction accuracy on all splits; errors occur on the first turn before flows are injected and predictions are random (see Example \ref{example_output/convo_6601_large} in Appendix~\ref{appendix:example_dialogues}). 

Without flow information (i.e. +$\emptyset$ and +L), flow accuracy drop dramatically on the generalization splits. 
For +$\emptyset$, it is \quasi0\%: as expected, %
the model cannot predict unknown flow names. However, unknown flows \textit{can} be predicted with non-zero accuracy (\quasi20-27\%) by providing the list of legal flow names (+L). %

\textit{Flow prefix} accuracy is significantly higher. %
With the list of flows (+L), it is \quasi57\% for Split 1 %
and \quasi46\% %
for Split 3 when even the \textit{flow prefix itself} is unseen during training.  Without the list of legal flows (+$\emptyset$), flow prefix prediction decreases from \quasi41\% to \quasi17\% from Splits 1 to 3. This suggests that other information from the context is being used to infer a reasonable flow prefix.

\subsection{Specific flow errors}
\label{sec:flow_errors}

Analyzing the errors of our model we observe that for the +$\emptyset$ variant, many of the errors seem reasonable. 
All of the \code{search-faq} type flows (\code{boots}, \code{short}, \code{jeans}, \code{jacket}) are largely confused with \code{pricing} which shares the same first and third actions and also searches the Database. Likewise, \code{mistimed\_billing\_never\_bought} is often confused with \code{status\_mystery\_fee} which has a semantically similar name and differs only in that it requires identity verification instead of purchase validation.
Others also seem reasonable based on the flow name, but actually do not share most actions 
and have very different purposes, like \code{promo\_code\_out\_of\_date} and \code{timing}, \code{missing} and \code{status\_delivery\_time}, \code{bad\_price\_competitor} and \code{pricing}. %
We provide flow prediction confusion matrices in Figures \ref{fig:flow_conf_matrix_large_d16-1_seed2_split3} and \ref{fig:flow_conf_matrix_large_d16-1a_seed2_split3} of the Appendix~\ref{sec:flow-prediction-confusion-matrices}.

Our models also predict totally new flows, and we have observed them create flows such as: \code{warranty}, \code{status\_credit\_rejected}, \code{manage\_change\_password}, \code{status\_stock}, and \code{manage\_change\_name}. %
These appear to be combinations of existing flow names.

For +L, when the list of flows is provided, %
all of the above errors also occur. The percentage of predicted flows in the set that was seen in training, test, or neither for both +L and +$\emptyset$ are shown in Table \ref{tab:flow_from_where}. Providing a list of allowable flows greatly increases the number of unseen flows predicted at test time. Future exploration could focus on how to encourage the model to rely less on training flows. %

\begin{table}[]
    \centering
    \resizebox{0.48\textwidth}{!}{
    \begin{tabular}{lccc}
        \hline
        Model configuration & \% train & \% test only\ & \% neither  \\
        \hline
        T5-large finetuned & 90.1 & 0.01 & 9.4 \\
        + L & 59.2 & 32.1 & 8.7 \\
        \hline
    \end{tabular}
    }
    \caption{Source of predicted flows for T5-large models on Split 3 (averaged over 3 seeds)}
    \label{tab:flow_from_where}
\end{table}

\section{Conclusion}
By training T5 text2text models for task-oriented dialogue and augmenting the dialogue context with plan information, we show that large language models can and do make use of provided sequences of action plans and are able to generalize to new action sequences, flows, and actions beyond those seen in training.

An advantage of our framework is that we could obtain workflow prompts from a symbolic planning mechanism or other types of external API calls. 
Symbolic planner correctness guarantees
enable a maintainable system, allowing new workflows to be added, and intermediate steps to be changed and re-assembled, that could in turn be used for generate novel prompts.
Appendix \ref{sec:planning-symbolic} shows how a set of workflows can be formulated as symbolic STRIPS planning~\cite{fikes-nilsson-aij1971}, and how adding an additional slot to an action automatically adjusts existing workflows, %
showing how symbolic planning is a promising direction for intentional modification of workflows. 
Moreover, symbolic planning methods could also be used to dynamically update plans mid-dialogue %
generating the equivalent of a personalized %
dialogue trees \cite{muise2019planning}.

This work is a first step towards our goal of integrating plans with LLMs to improve generalization in multi-step task-oriented dialogue.

\section*{Limitations}
There are a few key limitations with these experiments. Firstly, we provide the ground-truth oracle flow to the models, thus ensuring correct and useful action plans, but in practice, these would only be available by direct confirmation with the user. Thus, better flow prediction is necessary. We might improve flow prediction by  a) further augmenting the context with descriptions of the flows or their associated action sequences, thus rendering explicit the connection between flow names and goals, b) only predict the flow once enough information is provided in the dialogue and c) ensure that in training we provide examples of recovery from an incorrect flow prediction, which would allow the conversation to shift as more information becomes available. The latter would likely involve data augmentation and adding noise to training. See Appendix~\ref{sec:flow-prediction-accuracy-across-turns} for an analysis of flow performance across dialogue turns, showing that indeed flow and flow prefix prediction improves as more information is gathered, but does not plateau to perfect prediction.

Secondly, there is no connection between actions and the slots they require in our baseline experiments, which is problematic for unseen actions. %
As with flow information, slot types could be provided in the context as part of an ontology. Preliminary explorations along these lines are provided in Appendix~\ref{sec:slot_metrics}.
Alternatively, this could be provided immediately after the model has made an action prediction for the predicted action only. However, a mapping between the target slot-values and slot-types does not exist in ABCD. We intend to attempt to reconstruct these slot-types and, if possible, produce and make use of a subset of the ABCD dataset augmented with this information, since reconstruction will not be perfect. See section~\ref{sec:planning-detail} for the slots that can be supported by the symbolic planning model. 

Thirdly, the plans in ABCD are relatively short and simple. There are also no side effects for executing incorrect actions and no need to "backtrack." This reduces the utility of a full-scale planner that would be important in a more complex setting. We are not aware of a task-oriented dialogue dataset that contains such complexity. Text-worlds, intended for training reinforcement learning models, may contain the necessary world-state complexity, but often do not contain sufficient language complexity or natural interactions.

Fourthly, the plans that we provide to ABCD follow the provided ontology, but do not capture all of the information in the text-based descriptions provided in the guidelines to the WoZ. Some workflows contain "if-then" type statements which are not encoded in the raw action sequences in the ontology. There is also no way to "guess" the outcome of the ask-the-oracle action (a random output generator) based on the conversation context alone. Preprocessing the dataset with "lookahead" information would be necessary to fully arm the model. This would likely produce a dramatic improvement in the generalization splits.

Another limitation lies in our choices of a symbolic model. Our actions are deterministic while \cite{muise2019planning} consider actions that could lead to more than one possible effects. They focus on pre-computing precise plans while we focus on improving LM generalization. If during dialogue, an action has an unexpected effect or the state changes, a common approach is to replan that is safe as far as it is possible to recover and finish the dialogue. Evaluating this setting would require us to evaluate the LM in an interactive fashion, distinguishing physical actions like emitting an utterance of applying an action from search exploring the hypothetical space of future dialogues. Such evaluations are common when using symbolic planning in  robotics and other grounded areas of AI.

We also discarded evaluating our hypothesis using the popular task-oriented dialogue datasets MultiWoz \cite{budzianowski_multiwoz_2020}
and SGD \cite{rastogi_towards_2020}.
In principle, SGD schema specify the slots resulting from some intents, that are in turn used by other intents. 
However both datasets confound in \textit{intent} \textit{what to do} and \textit{how to solve it}.
Indeed, in both cases the data generation associates utterances to one action, so the main challenge is identifying which action to do and filling the corresponding slots.
The importance of novel sequences of actions is harder to observe in this case.
The simplified linear workflows of the ABCD dataset could be expressed as a SGD schema, hinting that prompting with symbolic plans would be relevant for cases where intent is associated with slots or paths in the schema (see section \ref{sec:planning-dialogues}).

We also do not explore or provide any mechanisms for identifying or recovering from miscommunication, or adapting to changing goals, both of which could also benefit from an explicit planning-based approach. Nor do we incorporate constraints on decoding, something which would be important in an actual application and likely provide a considerable boost to performance.

It is possible to augment dialogue context with the plan (+P) by explicitly asking the user to confirm their intent, through the use of a classifier, from predictions made by the model itself, or a combination of predictions and confirmations. However, we conjectured that automated prediction performance will be limited at the start of the dialogue. In Appendix ~\ref{sec:flow-prediction-accuracy-across-turns} we confirm that prediction performance over dialogue turns is indeed initially low, but it increases rapidly after the first turns. This suggests that it might be possible to use fully automated flow prediction to produce appropriate action-plans.

\section*{Ethics Statement}

Dialogue systems, if deployed in practice, must consider the impact they can have on users.  We have performed no bias correction, analysis, or filtering of model outputs. Thus, any real system that uses such a LM-based model should at minimum be analyzed for harmful utterances and differing performance across possible user-groups and likely have strong filters on their outputs.

Similarly, a dialogue system used for executing business workflows has the potential of causing negative impacts on a user's account if it makes errors or performs unwanted actions, to the detriment of either the user or the company. As in normal software development, rigorous testing of systems should be performed before deployment.

Such systems are also often intended to replace human support agents, which, if successfully and widely deployed could have a negative impact on workers if re-skilling or re-allocation is not implemented. However, in many cases such systems still require human oversight and can serve to enhance the productivity of workers, allowing them to focus on more interesting and challenging tasks. 

Lastly, the focus of this work on large language models may also have negative effects. Such models are increasing in size and being controlled more and more by large corporate entities and are requiring ever greater compute. This reduces access to research requiring them, and promoting and using such models supports this trend, especially if it leads to the exclusion of other types of research.

\bibliography{anthology,0mainbib}
\bibliographystyle{acl_natbib}

\appendix

\section{Related Work (extended)}
\label{sect:related-extended}

\subsection{Large language models} The development and public availability of large pretrained transformer-based language models has revolutionized the field of NLP, producing state of the art results across all tasks, including dialogue. Ever-larger models continue to provide improved results on these tasks, although there is debate around whether there is a limit to improvements that can be achieved in this fashion. There are a number of flavours of these models, all based around a Transformer~\cite{vaswani_attention_2017} backbone: encoder-only models, like BERT~\cite{devlin_bert_2018}; decoder only models like GPT-2/3~\cite{raffel_exploring_2020, brown_language_2020}; and encoder-decoder models, like T5~\cite{raffel_exploring_2020} and BART~\cite{lewis_bart_2019}. Given the massive corpora used to pretrain these models, their performance on zero-shot and few-shot tasks is very good~\cite{raffel_exploring_2020, brown_language_2020} and a new paradigm for performing few-shot or zero-shot generalization has emerged: text-based prompting and reframing tasks in a text2text format. The basic idea is to provide extra information in the input context to the model that will allow it to perform the task. This could include a description of the task, background information, or a few examples. This theoretically allows the model to generalize to any new task without the training of an independent classification layer. This text2text paradigm can be used with or without finetuning on the downstream dataset or task.

\subsection{Task-oriented dialogue datasets}

Task-oriented datasets are typically organized around slot-filling tasks or api calls and differ in terms of the number of domains/intents that they cover and the number of slot types, but also in various extra annotations. Newer datasets like SGD aim to increase the number of domains and api calls and provide free-form text descriptions of the various slots and intents. MultiWoz contains explicit "action prediction" labels, which are a structured prediction of the utterance that the model will generate (as opposed to a structured form of the necessary db call), and "belief state tracking" labels indicating the state of known information or slots required for the database calls. The ABCD dataset was created to provide more complex intents and api calls than existing datasets, along with restrictions to which solution can be provided according to a set of business rules provided only to the agent. The concept of flows, or sequences of actions, that need to be accomplished in ABCD in order to resolve a problem brings this dataset in some ways closer to planning and problem solving datasets than traditional slot-filling ones. Although it is not extensively used in the literature yet, we use ABCD for our experiments in order to test our hypotheses in a more complex environment. %

\subsection{Planning} Another approach to dialogue involves the explicit use of symbolic planners \cite{geffner2013concise}. Dialogue planning can be seen as a combination of planning with complete information and sensing of the environment and changes therein~\cite{steedman_planning_2007}. Dialogue planning involves the two separate but related tasks of discourse planning: low-level selection of appropriate dialogue acts and domain/acting planning~\cite{steedman_planning_2007}. In section \ref{sec:planning-dialogues} we should to how to build partially on that line of work.

Symbolic reasoning methods excel at 
inferring logically consistent conclusions from facts and rules, which can be chained together to form provably correct plans.
However, their traditional use is confined to cases where problems are %
formulated symbolically. Automated planning \cite{geffner2013concise} has been applied to solve two distinct tasks for task-oriented dialogue: %
 \textit{discourse planning} (the low-level selection of dialogue acts) and \textit{domain/action planning}~\cite{steedman_planning_2007, muise2019planning}. %

In terms of using planning in a dialogue system, the work of 
\cite{muise2019planning} is perhaps the closest to our own if we had decided to run our experiments using the symbolic planning discussed in Appendix~\ref{sec:planning-symbolic}
We also use the precondition structure of sequences of actions to dynamically generate plans. However, there are many differences between our approaches. First, our symbolic planning model is only concerned with the composition of domain actions or API calls. Second, in our approach symbolic plans are used only as conditioning for a large neural language model which in turn produces utterances and actions to be executed. Third, in our exploration we rely on pre-computed plans. While our framework allows plans to be dynamically determined on-the-fly, this functionality is simulated in our experiments by holding out workflows and action sequences in our data set splits. Finally, our approach enables zero-shot generalization without compiling into a tree or graph or all possible behaviours.

Other work uses planning/RL in text-based games~\cite{yao_keep_2020, micheli_language_2021}, which are dialogue-like and involve more complex interactions with the environment and state changes than most task-oriented dialogue datasets, though usually less fluent and complex language. 
Our work differs from \cite{yao_keep_2020} as their notion of state and reasoning are more primitive. For instance, they consider masking out actions for expressing business rules, but that only scratches the surface of a fundamental problem of sequential decision making: it is hard to anticipate the consequences of a current decision. In contrast, symbolic planning build plans that achieve the goal, offering more robustness.

Some other work incorporates certain aspects of planning, such as \citet{gao_robust_2020} who encode explicit business rules into a filter applied to their action prediction (and use a few other extra modules like an explicit NER and semantic parser for producing kb queries), but otherwise use a standard text2text setup, or such as \citet{xie_converse_2022} who incorporate a tree-structure for dialogue policy management within a standard task-oriented dialogue pipeline. While they might consider new trees in new situations, our line of work is compatible with short  declarations of changes or adjustments to the logic of the dialogue. 

In our experiments, we generate explicit sequential plans for the model to follow and incorporate this into our text2text dialogue setup and apply this to the complex language of a task-oriented dialogue dataset, ABCD.

\subsection{Evaluation Methods} The evaluation of dialogue systems is an open challenge. Utterance quality, judged by naturalness or relevance, cannot be effectively judged with standard automated metrics that rely on word-overlap or perplexity~\cite{liu_how_2016}. This makes evaluating systems beyond strict adherence to the labels provided in dialogue datasets very difficult. However, metrics such as BLEU and ROUGE are still the standard method for evaluating utterance quality in dialogue~\cite{lin_bitod_2021, thoppilan_lamda_2022, hosseini-asl_simple_2020, wu_dg2_2021}. These are sometimes supplemented by human-evaluation of quality measures, such as sensibleness, specificity, interestingness, fluency and factual accuracy~\cite{thoppilan_lamda_2022, ham_end--end_2020}, although this is very expensive and prohibitive at a large scale. Most systems rely on the annotations provided by datasets such as MultiWoz (e.g. ~\cite{he_galaxy_2022, hosseini-asl_simple_2020, ham_end--end_2020}), which contain explicit information about correct slot-values and intents, but do not encode multiple methods of solving a problem, if such methods exists. %

\section{Symbolic Planning and Dialogues}
\label{sec:planning-symbolic}

We begin here by formally defining planning problems, then discuss their encoding as STRIPS problems using the PDDL programming language. 

As we discussed above, we have implemented all our workflows in PDDL allowing them to be dynamically generated if desired; however, for our experiments above we have used a (faster) lookup table. The planning framework therefore offers a practical advantage in terms of maintainability of a system using our approach.

An action theory is a tuple $\Pi = \langle \Props, \Ops \rangle$,
where $\Props$ is a set of fluent or proposition symbols, and $\Ops$ is a set of operators or actions.
$\Facts$ the set of possible facts, where
a fact is an assignments of fluent $p$ in $\Props$ to true or false.
In this paper we use facts to describe real or hypothetical context or states.
A subset of facts from $\Facts$ is \textit{consistent} when it contains no fluent assigned to both true and false. 
Every operator $O \in \mathcal{O}$ is associated with its preconditions $\Opre{O} \subseteq \Facts$,
and its effect $\Oeff{O} \subseteq \Facts$.
Both $\Opre{O}$ and $\Oeff{O}$ are assumed to be consistent.

A state $s \subseteq \Facts$ is a set of assignments of the fluents $\Props$, such that all the fluents in $\Props$ are assigned in $s$. 
An action $a$ is {applicable} in a state $s$ if
$\Opre{a} \subseteq s$.  The {resulting state} of applying an applicable action $a$ in a
state $s$ is the state $f(s,a) = (s \setminus \Oneff{a}) \cup \Oeff{a}$, 
where $\Oneff{O}$ assigns to each fact in $\Oeff{O}$ its opposite value.
An action theory with an initial state and set of possible goal states is known as a
STRIPS planning problem or task ~\cite{fikes-nilsson-aij1971}. A solution to such problem is a plan or sequence of actions that is guaranteed to achieve the goal.
Planners receives a new STRIPS problems and attempt to return a plan without requiring any training.
In principle, a STRIPS problem might have no plan, or it might be hard to obtain as STRIPS plan existence is NP-had.
However, state-of-the-art planners can obtain plans with hundreds or thousand in seconds for a problem they have no seen before
\cite{geffner2013concise}. %

One can then encode STRIPS problems using the Planning Domain Definition Language (PDDL), and obtain plans by executing the planner. 
PDDL allows one to define STRIPS problems in a more compact way by using objects. The facts and actions are all possible instantiations of predicates and actions schemas over a fixed set of objects ~\cite{mcdermott-et-al-tr1998}.

\subsection{Planning Dialogues}
\label{sec:planning-dialogues}

To apply planning to dialogue, we consider propositions $\Props$ (statements with a true/false value) about the \textit{state} of the dialogue. These include slot value assignments, whether a workflow/goal has been achieved, and the order in which actions must appear in a given workflow. 
We then define the preconditions $\Opre{a}$ of an action as the slots which must be (or not be) available in the state to execute it, and the $\Oeff{a}$ as the changes it produces on the available slots. Actions are thus expressed as functions with a fixed number of arguments: $a_t = a^{a}(a^s_1,a^s_2, \ldots, ,a^s_K)$. 
This enables simple slot-filling setups %
but can also express more complex dialogues.

Given an initial state describing the slots already available in the context, the goal of a user intent, and an action theory $\Pi$ describing how actions depend on and affect slots,
we can formulate a STRIPS planning problem.
The set of all possible \textit{plans} corresponds to the set of all possible dialogue paths, taking actions from the initial state to the goal. For a given \textit{workflow}, the \textit{plan} updates at each stage of the dialogue to reflect the remaining \textit{actions} needed to complete the workflow. 

By holding out new pre-computed plans from training in our experiments below, we simulate the effect of new constraints, rules and tasks being added to the planning mechanism. The resulting plans will contain unseen actions, and sequences of actions. Since the planner is symbolic, there are no errors in such plans, except for the mismatch between the plans as specified in the knowledge base and justified deviations from that plan. We examine how an LLM conditioned on these plans may better adapt to performing new tasks in a zero-shot setting.

\subsection{Details on STRIPS formulation for ABCD dialogue}
\label{sec:planning-detail}

Consider the simplified case where there a few possible actions $a, b, c, d$, and there are ony two workflows. The workflow $w_1$ solved by executing the sequence of actions $\langle c, a\rangle$ while
$w_2$ solved by executing the sequence of actions $\langle c, d, b\rangle$. 
Let us assume such actions use not slots. Let us describe the possible behaviours using an STRIPS planning problem. We will need internal tokens to make sure we are achieving the goal.

This is an initial but incomplete description action theory for this situation.

\begin{itemize}
    \item Propositions $\Props$: \code{did-a}, \code{did-b}, \code{did-c}, \code{did-d}, \code{finished-w-1}, \code{finished-w-2}
    \item For each action $X$ in $\{ a, b, c, d \}$, an operator in  $\Ops$: \code{action-$X$} with effect \code{did-$X$}=true.
\end{itemize}

This action theory allows to execute any action at any moment, while the fluents \code{finished-w-$k$} remain unchanged. One improvement would be to make sure that \code{finished-w-1} is only added after the actions \code{action-$c$} and \code{action-$a$} are executed in that order. 
Another improvement is to add slots to each action, and to add actions that simulates when the language model provides them.

We wrote scripts translating the workflows of the ABCD dataset into full STRIPS problem written in PDDL syntax. Setting recover\_username as the flow to be completed in goal, calling a planner returned the following plan. Each line is an STRIPS action. The actions contain spaces because of how PDDL generates a set of possible actions. We added lines starting with ";;" to comment parts of the plan

\begin{lstlisting}[style=quote_style, language={}, firstnumber=1]
;; 4 slots needed for pull-up-account
get-slot slot_shipping_option
get-slot slot_payment_method
get-slot slot_customer_name
get-slot slot_account_id
;; action that checks pull-up-account as all the slots. This enables flexible slots. 
complete-button-slot button_pull-up-account slot_customer_name slot_account_id slot_shipping_option slot_payment_method
;; action pull-up-account is executed
do action_pull-up-account
;; slots for verify-identity
get-slot slot_order_id
get-slot slot_zip_code
complete-button-slot button_verify-identity slot_customer_name slot_account_id slot_order_id slot_zip_code
do action_verify-identity
;; We have set the flow is recover_username. This action starts verification
choose-flow flow_recover_username
;; next-step-flow actions control that actions for a flow are executed in order
next-step-flow flow_recover_username s_0 s_1 button_pull-up-account
next-step-flow flow_recover_username s_1 s_2 button_verify-identity
;; last actions of flow was executed so flow will be complete
complete-flow flow_recover_username s_2
\end{lstlisting}

The plan above includes the verification that all the slots are being provided, and the precise  sequence of steps is followed. 

For augmenting the context, sec~\ref{sec:augmenting_context}, the plan above would be stripped into the lines starting with \code{do action\_}:

\begin{lstlisting}[style=quote_style, language={}, firstnumber=1]
do action_pull-up-account
do action_verify-identity
\end{lstlisting}

The prompt is that text removing occurrences of \code{do action\_}.
If we want to use a prompt with slots, we could use:

\begin{lstlisting}[style=quote_style, language={}, firstnumber=1]
get-slot slot_shipping_option
get-slot slot_payment_method
get-slot slot_customer_name
get-slot slot_account_id
do action_pull-up-account
get-slot slot_order_id
get-slot slot_zip_code
do action_verify-identity
\end{lstlisting}

Let us suppose that there was a security incident in a company implementing the workflows described in the ABCD dataset.
A decision is taken that the action \code{verify-identify} would need a another slot called \code{account-uncompromised}. Such a slot would be provided by another action called \code{extra-verification}.

In PDDL we only need to make three small changes:
\begin{itemize}
    \item Add new slot \code{account-uncompromised} as a new proposition.
    \item Action \code{verify-identify} has a new precondition \code{account-uncompromised}.
    \item New action \code{extra-verification} with effect \code{account-uncompromised} = true.
\end{itemize}

This small intentional change would affect 22 out of the 55 workflows in the ABCD dataset, adding one step before each occurrence of \code{verify-identify} in 
Table~\ref{fig:kb_new-action-verify-identity}.

\section{Dataset splits}
\label{appendix:dataset_splits}

The number of examples in each dataset split is provided in Table~\ref{table/n_turns_per_split}, while Table~\ref{tab:kb_split_all} describes the division of flows into train/test for each split.

\begin{table}[h!]%
    \centering
    \small
     \resizebox{0.5\textwidth}{!}{
    \begin{tabular}{l c c c c}
    \hline
        Split & Standard & Split 1 & Split 2 & Split 3 \\
        \hline
        Train examples & 105175 & 67521 & 60183 & 42142\\ 
        Test examples & 13152 & 4333 & 5448 & 7927 \\
        \hline
    \end{tabular}
    }
    \caption{Dataset split sizes.}
    \label{table/n_turns_per_split}
\end{table}

\begin{table*}%
    \centering
    \resizebox{1\textwidth}{!}{
\begin{tabular}{l!{\color{black}\vrule}l!{\color{black}\vrule}l}
\hline
Workflow prefix group & Workflow name           & \multicolumn{1}{l}{Action sequence}                                                                               \\
\hline
account\_access       & recover\_username       & pull-up-account, \textbf{extra-verification}, verify-identity                                                      \\
\arrayrulecolor{black}\hline
manage\_account       & manage\_change\_address & pull-up-account, record-reason, \textbf{extra-verification}, verify-identity, update-account                       \\
                      & manage\_change\_name    & pull-up-account, record-reason, \textbf{extra-verification}, verify-identity, update-account                       \\ 
\hline
                      & manage\_change\_phone   & pull-up-account, record-reason, \textbf{extra-verification}, verify-identity, update-account                       \\
                      & manage\_payment\_method & pull-up-account, record-reason, \textbf{extra-verification}, verify-identity, update-account                       \\ 
\hline
order\_issue          & status\_mystery\_fee    & pull-up-account, \textbf{extra-verification}, verify-identity, ask-the-oracle, membership, update-order            \\
                      & status\_delivery\_time  & pull-up-account, \textbf{extra-verification}, verify-identity, ask-the-oracle, shipping-status, update-order       \\ 
\hline
                      & status\_payment\_method & pull-up-account, \textbf{extra-verification}, verify-identity, shipping-status, update-order                       \\
                      & status\_quantity        & pull-up-account, \textbf{extra-verification}, verify-identity, ask-the-oracle, shipping-status, offer-refund       \\ 
\hline
                      & manage\_upgrade         & pull-up-account, \textbf{extra-verification}, verify-identity, shipping-status, membership, update-order           \\
                      & manage\_downgrade       & pull-up-account, \textbf{extra-verification}, verify-identity, shipping-status, membership, update-order           \\ 
\hline
                      & manage\_create          & pull-up-account, \textbf{extra-verification}, verify-identity, shipping-status, membership, make-purchase          \\
                      & manage\_cancel          & pull-up-account, \textbf{extra-verification}, verify-identity, shipping-status, membership, offer-refund           \\ 
\hline
purchase\_dispute     & bad\_price\_competitor  & pull-up-account, record-reason, \textbf{extra-verification}, verify-identity, promo-code                           \\
                      & bad\_price\_yesterday   & pull-up-account, record-reason, \textbf{extra-verification}, verify-identity, promo-code                           \\ 
\hline
shipping\_issue       & status                  & pull-up-account, \textbf{extra-verification}, verify-identity, validate-purchase, ask-the-oracle, update-order     \\
\arrayrulecolor{black}\hline
subscription\_inquiry & status\_active          & pull-up-account, \textbf{extra-verification}, verify-identity, subscription-status, send-link, enter-details       \\
                      & status\_due\_amount     & pull-up-account, \textbf{extra-verification}, verify-identity, subscription-status, send-link, enter-details       \\ 
\hline
                      & status\_due\_date       & pull-up-account, \textbf{extra-verification}, verify-identity, subscription-status, send-link, enter-details       \\
                      & manage\_pay\_bill       & pull-up-account, \textbf{extra-verification}, verify-identity, subscription-status, enter-details, update-account  \\ 
\hline
                      & manage\_extension       & pull-up-account, \textbf{extra-verification}, verify-identity, membership, update-account, enter-details           \\
                      & manage\_dispute\_bill   & pull-up-account, \textbf{extra-verification}, verify-identity, membership, ask-the-oracle, offer-refund            \\
\hline
\end{tabular}
    }
    \caption{Flows after adding new action for verifying account is uncompromised. Some flows not effected by change were not included. See original flows in Table \ref{tab:kb_split_all}.}
    \label{fig:kb_new-action-verify-identity}
\end{table*}

\begin{table*}%
    \centering
    \resizebox{0.98\textwidth}{!}{
    \begin{tabular}{p{0.3cm}!{\color{black}\vrule}p{0.3cm}!{\color{black}\vrule}p{0.3cm}!{\color{black}\vrule}l!{\color{black}\vrule}l!{\color{black}\vrule}l}
         \arrayrulecolor{black}\hline
         \multicolumn{3}{c}{Split test set} & Workflow prefix group & Workflow name & Action sequence \\
         1 & 2 & 3 & & & \\ 
          \arrayrulecolor{black}\hline
         && U & account\_access & recover\_username & pull-up-account, verify-identity \\
         && U && recover\_password & pull-up-account, enter-details, make-password \\
         \arrayrulecolor{lightgray}\hline
         U & U & U && reset\_2fa & pull-up-account, enter-details, send-link \\
         \arrayrulecolor{black}\hline
         &&& manage\_account & status\_service\_added & pull-up-account, record-reason, ask-the-oracle, update-account, offer-refund \\
         &&&& status\_service\_removed & pull-up-account, record-reason, ask-the-oracle, update-account \\
         \arrayrulecolor{lightgray}\hline 
         U & U &&& status\_shipping\_question & pull-up-account, ask-the-oracle, send-link \\
         U & U &&& status\_credit\_missing & pull-up-account, record-reason, ask-the-oracle, promo-code \\
         \arrayrulecolor{lightgray}\hline 
         & U &&& manage\_change\_address & pull-up-account, record-reason, verify-identity, update-account \\
         & U &&& manage\_change\_name & pull-up-account, record-reason, verify-identity, update-account \\
         \arrayrulecolor{lightgray}\hline 
         U & U &&& manage\_change\_phone & pull-up-account, record-reason, verify-identity, update-account \\
         U & U &&& manage\_payment\_method & pull-up-account, record-reason, verify-identity, update-account \\
         \arrayrulecolor{black}\hline
         &&& order\_issue & status\_mystery\_fee & pull-up-account, verify-identity, ask-the-oracle, membership, update-order \\
         &&&& status\_delivery\_time & pull-up-account, verify-identity, ask-the-oracle, shipping-status, update-order \\
         \arrayrulecolor{lightgray}\hline 
         U & U &&& status\_payment\_method & pull-up-account, verify-identity, shipping-status, update-order \\
         U & U &&& status\_quantity & pull-up-account, verify-identity, ask-the-oracle, shipping-status, offer-refund \\
         \arrayrulecolor{lightgray}\hline 
         &&&& manage\_upgrade & pull-up-account, verify-identity, shipping-status, membership, update-order \\
         U &&&& manage\_downgrade & pull-up-account, verify-identity, shipping-status, membership, update-order \\
         \arrayrulecolor{lightgray}\hline 
         & U &&& manage\_create & pull-up-account, verify-identity, shipping-status, membership, make-purchase \\
         U & U &&& manage\_cancel & pull-up-account, verify-identity, shipping-status, membership, offer-refund \\
         \arrayrulecolor{black}\hline
         &&& product\_defect & refund\_initiate & pull-up-account, validate-purchase, record-reason, enter-details, offer-refund \\
         &&&& refund\_update & pull-up-account, validate-purchase, record-reason, offer-refund \\
         \arrayrulecolor{lightgray}\hline 
         U & U &&& refund\_status & pull-up-account, validate-purchase, notify-team, update-order \\
         & U &&& return\_stain & pull-up-account, validate-purchase, membership, enter-details, update-order \\
         \arrayrulecolor{lightgray}\hline 
         & U &&& return\_color & pull-up-account, validate-purchase, membership, enter-details, update-order \\
         U & U &&& return\_size & pull-up-account, validate-purchase, membership, enter-details, update-order \\
         \arrayrulecolor{black}\hline
         U && U & purchase\_dispute & bad\_price\_competitor & pull-up-account, record-reason, verify-identity, promo-code \\
         && U && bad\_price\_yesterday & pull-up-account, record-reason, verify-identity, promo-code \\
         \arrayrulecolor{lightgray}\hline 
         && U && out\_of\_stock\_general & pull-up-account, notify-team, promo-code \\
         U & U & U && out\_of\_stock\_one\_item & pull-up-account, record-reason, notify-team, make-purchase \\
         \arrayrulecolor{lightgray}\hline 
         & U & U && promo\_code\_invalid & pull-up-account, ask-the-oracle, membership, promo-code \\
         & U & U && promo\_code\_out\_of\_date & pull-up-account, ask-the-oracle, membership, promo-code \\
         \arrayrulecolor{lightgray}\hline 
         && U && mistimed\_billing\_already\_returned & pull-up-account, validate-purchase, record-reason, membership, update-order \\
         & U & U && mistimed\_billing\_never\_bought & pull-up-account, validate-purchase, ask-the-oracle, membership, update-order \\
         \arrayrulecolor{black}\hline
         && U & shipping\_issue & status & pull-up-account, verify-identity, validate-purchase, ask-the-oracle, update-order \\
         && U && manage & pull-up-account, shipping-status, validate-purchase, update-order \\
         \arrayrulecolor{lightgray}\hline 
         U & U & U && missing & pull-up-account, validate-purchase, record-reason, update-order, make-purchase \\
         U & U & U && cost & pull-up-account, validate-purchase, shipping-status, update-order, offer-refund \\
         \arrayrulecolor{lightgray}\hline 
         \arrayrulecolor{black}\hline
         && U & single\_item\_query & boots & search-faq, search-boots, select-faq \\
         && U && shirt & search-faq, search-shirt, select-faq \\
         \arrayrulecolor{lightgray}\hline 
         && U && jeans & search-faq, search-jeans, select-faq \\
         && U && jacket & search-faq, search-jacket, select-faq \\
         \arrayrulecolor{black}\hline
         &&& storewide\_query & pricing & search-faq, search-pricing, select-faq \\
         &&&& membership & search-faq, search-membership, select-faq \\
         \arrayrulecolor{lightgray}\hline 
         &&&& timing & search-faq, search-timing, select-faq \\
         &&&& policy & search-faq, search-policy, select-faq \\
         \arrayrulecolor{black}\hline
         && U & subscription\_inquiry & status\_active & pull-up-account, verify-identity, subscription-status, send-link, enter-details \\
         && U && status\_due\_amount & pull-up-account, verify-identity, subscription-status, send-link, enter-details \\
         \arrayrulecolor{lightgray}\hline
         U && U && status\_due\_date & pull-up-account, verify-identity, subscription-status, send-link, enter-details \\
         & U & U && manage\_pay\_bill & pull-up-account, verify-identity, subscription-status, enter-details, update-account \\
         \arrayrulecolor{lightgray}\hline 
         & U & U && manage\_extension & pull-up-account, verify-identity, membership, update-account, enter-details \\
         U & U & U && manage\_dispute\_bill & pull-up-account, verify-identity, membership, ask-the-oracle, offer-refund \\
         \arrayrulecolor{black}\hline
         && U & troubleshoot\_site & credit\_card & try-again, log-out-in, enter-details, make-purchase \\
         U & U & U && shopping\_cart & try-again, log-out-in, enter-details, make-purchase \\
         \arrayrulecolor{lightgray}\hline 
         & U & U && search\_results & try-again, log-out-in, make-purchase, instructions, notify-team \\
         U && U && slow\_speed & try-again, log-out-in, make-purchase, instructions, notify-team \\
         \arrayrulecolor{black}\hline
    \end{tabular}
    }
    \caption{Flows and associated action sequences split into train/test (seen/unseen) for dataset Splits 1, 2, and 3. Flows that are unseen in training are marked with a U.}
    \label{tab:kb_split_all}
\end{table*}

\section{Examples of predictions for dialogues}
\label{appendix:example_dialogues}
Section~\ref{sec:example-unseen-action-correct} shows  an example where
an unseen action is correctly executed by the agent, the name of which is drawn from the provided plan. %
Section~\ref{sec:example-woz-does-not-follow-flow} shows another case where the agent follows the plan for an unseen flow and action sequence. 
In both cases, we comment on the behaviour using T5-base and T5-large.

\subsection{Example: Unseen action correctly implemented by the agent}
\label{sec:example-unseen-action-correct}
\subsubsection{T5-base}
The example shown in Table \ref{example_output/convo_6601} contains a dialogue for the flow \code{recover-password} which contains the action \code{make-password} that is unseen in training. Though there are some small errors, the dialogue proceeds relatively smoothly.
On turn 1, the agent incorrectly predicts the flow to be \code{status\_service\_removed} given an empty context. 
After correctly executing \code{pull-up-account} on turn 3, on turn 4 the agent immediately attempts to execute the next action suggested by the updated plan \code{enter-details}. According to the ontology, \code{enter-details} requires only one of the following slots: \code{details, full\_address, account\_id, username, email, pin\_number, security\_answer}. However, in the training set, it is executed at least 1 time by the WoZ with a full name\footnote{As detected by the following regex to detect 2-word values: "\textbackslash s?\textbackslash S+\textbackslash s\textbackslash S+\textbackslash s?\$".}. This suggests that 1) the agent prioritizes executing the plan, 2) may not have a clear notion of which slots are required to execute a given action. 
After the WoZ requests the customer's username on turn 5, the agent correctly uses this information to execute \code{enter-details}. Recall that there is no duplicate execution of enter-details, since the agents' previous actions are not iteratively incorporated into the context. 
The agent successfully accomplishes the \code{make-password} action on turn 7, which is the only action in the updated action plan (see the changes in the context). It is possible that the agent guessed that the \code{make-password} action does not contain any slots because either 1) there are other actions without slots and this was a safe bet or 2) no new potential slot information has been provided by the user at this point.
On turn 8, agent correctly explains to the user that "your password has correctly been recovered", a phrase presumably compiled from the flow information and other similar phrases in the dialogues. However, the WoZ, as instructed by the guidelines, gives the customer the new password, while the agent does not know to do so. Note also that this turn is an example of a reasonable utterance provided by the agent, but one which has a very low rouge score, confirming that rouge is a poor metric for dialogue quality.

\subsubsection{T5-large}
The same example for the flow \code{recover-password} is shown in Table \ref{example_output/convo_6601_large} for T5-large. 
The T5-large agent does not attempt to immediately execute the \code{enter-details} action on turn 4 without first asking for more information. The information requested from the user on turn 4 is the account id and order id: the former is a legal slot for enter-details, while the latter is not. This suggests that an explicit connection between actions and their legal slots would be helpful.
The agent then makes two mistakes on turn 7. Firstly, it predicts the "action" prefix, but then proceeds to provide an utterance. Secondly, it requests that the user provide a new password, instead of having it be generated by the system. This is perhaps a reasonable error; a request is indeed being made to "make-password." 
On turn 8, the agent does inform the user that a new password has been created, but does not provide it's value, which is reasonable, since it appears nowhere in the provided context and is infact made up by the WoZ. As with T5-base, the comment on turn 9 is less helpful than the one provided by the WoZ.

\subsection{Example: WoZ does not follow labelled flow, but agent sticks to the plan}
\label{sec:example-woz-does-not-follow-flow}

\subsubsection{T5-base}
In the next example, shown in Table \ref{example_output/convo_2049}, the customer states that they cannot access their account because they lost their phone. According to the labels in the dataset, the agent should execute the \code{recover\_2fa} flow, which consists of the actions \code{pull-up-account, enter-details, send-link}. However, on turn 5, the WoZ seems to misunderstand that the user has lost their password and proceeds to execute the \code{make\_password} flow which consists of the actions they actually take \code{pull-up-account, verify-identity, make-password}. The customer never corrects the agent, though presumably their instructions tell them that they need to reset their 2fa (instructions provided to the customer are not provided in ABCD). This misunderstanding might be understandable: likely the customer was only provided with a phone number and is not familliar with the internal company procedures  Thus, the flow label associated with this example is incorrect and the corresponding plan that we have passed to the model is also incorrect. 

Despite the incorrect label, this example shows how the models react when provided with a plan. Note that the \code{make-password} action has never been seen in training, and the \code{reset\_2fa} plan and sequence of steps is also unseen. 

Firstly, the randomly predicted flow on turn 1 is incorrect.
On turn 2, the agent's comment is generic compared to the WoZ, but reasonable given that the provided action plan begins with \code{pull-up-account}: it asks for a full name, while the WoZ has already misdirected the customer. 
Turn 5 shows the model favouring the same generic utterance "I can help you with that" from turn 3, although note that this is not repetition because it was never incorporated into the context. The model also again asks for a name. 
Turn 8 has the model correctly providing acknowledgement to the user after executing the action on turn 7. 

Things start to fall apart on turn 10, when the next suggested action in the action plan is \code{enter-details}. According to the ontology, \code{enter-details} requires only one of the following slots: \code{details, full\_address, account\_id, username, email, pin\_number, security\_answer}. The agent immediately attempts to execute this action with a phone number, which, while incorrect, does appear in the training set \quasi3/546 times\footnote{According to a quick regex of "\textbackslash s?\textbackslash(?\textbackslash d\textbackslash d\textbackslash d\textbackslash)?\textbackslash s?\textbackslash d\textbackslash d\textbackslash d\textbackslash s?\textbackslash-?\textbackslash d\textbackslash d\textbackslash d\textbackslash d\textbackslash s?" which also detects one non-phone numbers and retrieves 4 turns.}. However, the WoZ never actually performs this action, so the agent continues to attempt to complete on almost every subsequent turn, with various other pieces of information that are provided by the user, only providing an acknowledgement response on turn 12, and successfully ending the conversation after the customer indicates that they do not need anything else. 

This conversation shows that the agent 1) is using the plan to guide the conversation and 2) is committed to accomplishing the actions provided in the plan, 3) does not have a clear notion of which slots should be provided to a given action. 

\subsubsection{T5-large}
The same example in which the WoZ does not follow the prescribed flow as executed by the T5-large agent is shown in Table \ref{example_output/convo_2049_large}. The agent's utterance on turn 2 does convey sympathy for the user's lost phone, something which is done by the WoZ, but not by the T5-base agent which provides a more generic response. However, the agent neglects to request for any information from the user. This is not incorrect in general, but would require the model to generate a follow-up utterance in which the information is requested, something that is not permissible in the current evaluation setup since we are evaluating turn-by-turn, resetting the context, and not incorporating the models' outputs. Similarly, the agent's utterance on turn 5 better acknowledges the user's question, but does not request the necessary information for executing the \code{pull-up-account} action. 

As with the T5-base model, the agent attempts to execute the \code{enter-details} action on turn 10 and 11, in accordance with the plan, although the WoZ is no longer executing the actions of the correct workflow. The agent continues to execute \code{enter-details} actions throughout the conversation with the new information being provided in the real conversation context. Unlike the T5-base model, however, it also attempts to execute \code{verify-identity} which is not in the plan, but is a common action to follow \code{pull-up-account} and the one executed by the WoZ. The agent does not execute the final \code{send-link} action.

This example suggests that a connection between actions and their slots would be beneficial, but also that some form of constraints on the model's outputs could keep conversations on track. However, more importantly, it highlights issues with the ABCD dataset labels and suggests that further experimentation should more carefully prepare plans according to the actual WoZ conversation context instead of based on the possibly incorrect workflow labels.

\begin{table*}%
    \centering
    \footnotesize
    \begin{tabular}{p{0.1cm}p{0.5cm}p{4cm}p{4cm}p{6cm}}
    \hline
&& Expected & Actual & Context (truncated) \\
\hline\hline
       1&& flow: recover\_password; agent: Hello, how can i help you today & flow: \red{status\_service\_removed}; agent: Hello, how can I help you today\red{?} & ~ \\ \hline
2&\multicolumn{3}{l}{customer: Hi I forgot my password to my account. My name is Crystal Minh.} \\ 
\hline
3&        & flow: recover\_password; action: pull-up-account: crystal minh & flow: recover\_password; action: pull-up-account: crystal minh & agent: Hello..., flow: recover\_password; action\_plan: pull-up-account, enter-details, make-password;  \\ 
        \hline
4&        & flow: recover\_password; agent: Okay, could i get your username please & flow: recover\_password; \red{action: enter-details: crystal minh} & agent: Hello... action: pull-up-account: crystal minh flow: recover\_password; action\_plan: enter-details, make-password;  \\ 
        \hline
5&\multicolumn{3}{l}{customer: cm374950        } \\
\hline
6&        & flow: recover\_password; action: enter-details: cm374950 & flow: recover\_password; action: enter-details: cm374950 & agent: Hello,... customer: cm374950 flow: recover\_password; action\_plan: enter-details, make-password;  \\ 
        \hline
7&        & flow: recover\_password; action: make-password:  & flow: recover\_password; action: make-password: & agent: Hello, ... action: enter-details: cm374950 flow: recover\_password; action\_plan: make-password;  \\ 
        \hline
 8&       & flow: recover\_password; agent: Okay, here is your new password 3mihalbfbem & flow: recover\_password; agent: Okay, \red{your password has been recovered. Is there anything else i can help you with today?} & agent: Hello... action: make-password:  flow: recover\_password; action\_plan: ;  \\ 
        \hline
 9&       & flow: recover\_password; agent: You can log in and change it again if you want to. Is there anything else i can help you with & flow: recover\_password; agent: Is there anything else i can help you with today & agent: Hello... agent: Okay, here is your new password 3mihalbfbem flow: recover\_password; action\_plan: ;  \\ 
        \hline
10&\multicolumn{3}{l}{customer: great. thanks} \\ 
11&\multicolumn{3}{l}{customer: that's all} \\ 
\hline
12&        & flow: recover\_password; agent: Okay, have a nice day & flow: recover\_password; agent: \red{Have a great day} & agent: Hello... customer: great. thanks customer: that's all flow: recover\_password; action\_plan: ; \\ 
    \hline
    \end{tabular}
    \caption{Example dialogue with unseen action make-password. Split-3, D31-1, T5-base, seed: 0, conversation id: 6601. Incorrect predictions are in red.}
    \label{example_output/convo_6601}
\end{table*}
\begin{table*}%
    \centering
    \footnotesize
    \begin{tabular}{p{0.1cm}p{0.5cm}p{4cm}p{4cm}p{6cm}}
    \hline
&& Expected & Actual & Context (truncated) \\
\hline\hline
       1&& flow: recover\_password; agent: Hello, how can i help you today & flow: \red{status\_service\_removed}; agent: Hello, how can I help you today\red{?} & ~ \\ \hline
2&\multicolumn{3}{l}{customer: Hi I forgot my password to my account. My name is Crystal Minh.} \\ 
\hline
3&        & flow: recover\_password; action: pull-up-account: crystal minh & flow: recover\_password; action: pull-up-account: crystal minh & agent: Hello..., flow: recover\_password; action\_plan: pull-up-account, enter-details, make-password;  \\ 
        \hline

4&        & flow: recover\_password; agent: Okay, could i get your username please & flow: recover\_password; Okay, can i get your account id and order id please & agent: Hello... action: pull-up-account: crystal minh flow: recover\_password; action\_plan: enter-details, make-password;  \\ 
        \hline
        
5&\multicolumn{3}{l}{customer: cm374950        } \\
\hline
6&        & flow: recover\_password; action: enter-details: cm374950 & flow: recover\_password; action: enter-details: cm374950 & agent: Hello,... customer: cm374950 flow: recover\_password; action\_plan: enter-details, make-password;  \\ 
        \hline
        
7&        & flow: recover\_password; action: make-password:  & flow: recover\_password; action: \red{Okay, what is your password please} & agent: Hello, ... action: enter-details: cm374950 flow: recover\_password; action\_plan: make-password;  \\ 
        \hline
 8&       & flow: recover\_password; agent: Okay, here is your new password 3mihalbfbem & flow: recover\_password; agent: Okay, \red{i have created a new password for you.} & agent: Hello... action: make-password:  flow: recover\_password; action\_plan: ;  \\ 
        \hline
 9&       & flow: recover\_password; agent: You can log in and change it again if you want to. Is there anything else i can help you with & flow: recover\_password; agent: Is there anything else i can help you with & agent: Hello... agent: Okay, here is your new password 3mihalbfbem flow: recover\_password; action\_plan: ;  \\ 
        \hline
10&\multicolumn{3}{l}{customer: great. thanks} \\ 
11&\multicolumn{3}{l}{customer: that's all} \\ 
\hline
12&        & flow: recover\_password; agent: Okay, have a nice day & flow: recover\_password; agent: Okay, have a nice day & agent: Hello... customer: great. thanks customer: that's all flow: recover\_password; action\_plan: ; \\ 
    \hline
    \end{tabular}
    \caption{Example dialogue with unseen action make-password. Split-3, D31-1, T5-large, seed: 0, conversation id: 6601. Incorrect predictions are in red.}
    \label{example_output/convo_6601_large}
\end{table*}

\begin{table*}%
    \centering
    \footnotesize
    \begin{tabular}{p{0.1cm}p{0.5cm}p{4cm}p{4cm}p{6cm}}
    \hline
&& Expected & Actual & Context (truncated) \\
\hline\hline
        1&& flow: reset\_2fa; agent: Hi! Thank you for contacting Acme today. How may I help you? & flow: \red{status\_service\_removed}; agent: \red{Hello, how can I help you today?} & ~ \\ \hline
   2& \multicolumn{3}{l}{customer: yes, i can't access my account because i lost my phone }\\
    3&   & flow: reset\_2fa; agent: I am so sorry to hear you lost your phone.  Do you need to get your username and password or do you have one of them? & flow: reset\_2fa; agent: \red{I'll be glad to help you with that. May I have your full name please?} & agent: Hi! ... customer: yes, i can't access my account because i lost my phone flow: reset\_2fa; action\_plan: pull-up-account, enter-details, send-link;  \\ \hline
    4& \multicolumn{3}{l}{customer: can i give you my phone number instead }\\
     5&   & flow: reset\_2fa; agent: Can I first start with your full name? & flow: reset\_2fa; agent: \red{I can help you with that. May I have your} full name \red{or account ID please?} & agent: Hi! ... customer: can i give you my phone number instead flow: reset\_2fa; action\_plan: pull-up-account, enter-details, send-link;  \\ \hline
    6& \multicolumn{3}{l}{customer: Albert Sanders }\\
    7&  &  flow: reset\_2fa; action: pull-up-account: albert sanders & flow: reset\_2fa; action: pull-up-account: albert sanders & agent: Hi! ... customer: Albert Sanders flow: reset\_2fa; action\_plan: pull-up-account, enter-details, send-link;  \\ \hline
    8&   & flow: reset\_2fa; agent: Can you also provide me with your phone number and email address & flow: reset\_2fa; agent: \red{Thank you Albert. I was able to pull up your account.} & agent: Hi! ... action: pull-up-account: albert sanders flow: reset\_2fa; action\_plan: enter-details, send-link;  \\ \hline
    9& \multicolumn{3}{l}{customer: i don't have an email. the phone number is 330-822-4754 }\\
     10&  & flow: reset\_2fa; agent: Ok, that is fine.  What is your zip code & flow: reset\_2fa; \red{action: enter-details: 330-822-4754} & agent: Hi! ... customer: i don't have an email. the phone number is 330-822-4754 flow: reset\_2fa; action\_plan: enter-details, send-link;  \\ \hline
     11&  & flow: reset\_2fa; action: verify-identity: albert sanders, 69233, 330-822-4754 & flow: reset\_2fa; \red{action: enter-details: 69233} & agent: Hi! ... customer: 69233 flow: reset\_2fa; action\_plan: enter-details, send-link;  \\ \hline
      12& & flow: reset\_2fa; agent: Thank you for all the information.  Your username is asanders1 & flow: reset\_2fa; agent: Thank you, I was able to verify your identity. & agent: Hi! ... action: verify-identity: albert sanders, 69233, 330-822-4754 flow: reset\_2fa; action\_plan: enter-details, send-link;  \\ \hline
    13&\multicolumn{3}{l}{customer: ha..of course }\\
    14&   & flow: reset\_2fa; agent: I can not actually view your old password for security reason so I have to create a new one & flow: reset\_2fa; \red{action: enter-details: asanders1} & agent: Hi! T... customer: ha..of course flow: reset\_2fa; action\_plan: enter-details, send-link;  \\ \hline
    15&\multicolumn{3}{l}{customer: ok  }\\
     16&  & flow: reset\_2fa; agent: Can I have your account pin or the answer to your security question & flow: reset\_2fa; agent: \red{Is there anything else I can help you with today?} & agent: Hi! ... agent: I can not actually view your old password for security reason so I have to create a new one customer: ok flow: reset\_2fa; action\_plan: enter-details, send-link;  \\ \hline
    17&\multicolumn{3}{l}{customer: security question answer is Alexander }\\        
     18& &  flow: reset\_2fa; agent: Thank you. One moment while I generate the password & flow: reset\_2fa; \red{action: enter-details: asanders1} & agent: Hi! ...customer: security question answer is Alexander flow: reset\_2fa; action\_plan: enter-details, send-link;  \\ \hline
      19& & flow: reset\_2fa; action: make-password:  & flow: reset\_2fa; action: \red{enter-details: asanders1} & agent: Hi! ... agent: Thank you. One moment while I generate the password flow: reset\_2fa; action\_plan: enter-details, send-link;  \\ \hline
      20& & flow: reset\_2fa; agent: Your password is rox6fnwo33e & flow: reset\_2fa; \red{action: enter-details: asanders1} & agent: Hi! ... action: make-password:  flow: reset\_2fa; action\_plan: enter-details, send-link;  \\ \hline
    21&\multicolumn{3}{l}{customer: ok great}\\ 
    22&   & flow: reset\_2fa; agent: did you have any other questions today? & flow: reset\_2fa; action: \red{enter-details: rox6fnwo33e} & agent: Hi! ... customer: ok great flow: reset\_2fa; action\_plan: enter-details, send-link;  \\ \hline
    23&\multicolumn{3}{l}{customer: nope. thanks so much!} \\
     24&  & flow: reset\_2fa; agent: You're Welcome.  Have a great day & flow: reset\_2fa; agent: \red{Thank you so much for contacting Acme. Have a great day!} & agent: Hi! ...customer: nope. thanks so much! flow: reset\_2fa; action\_plan: enter-details, send-link; \\ \hline
    \end{tabular}
    \caption{Example dialogue with unseen action make-password. Split-3, D31-1, T5-base, seed: 0, conversation id: 2049. Incorrect predictions are in red.}
    \label{example_output/convo_2049}
\end{table*}

\begin{table*}%
    \centering
    \footnotesize
    \begin{tabular}{p{0.1cm}p{0.5cm}p{4cm}p{4cm}p{6cm}}
    \hline
&& Expected & Actual & Context (truncated) \\
\hline\hline
        1&& flow: reset\_2fa; agent: Hi! Thank you for contacting Acme today. How may I help you? & flow: \red{status\_service\_removed}; agent: \red{Hello, how can I help you today?} & ~ \\ \hline
   2& \multicolumn{3}{l}{customer: yes, i can't access my account because i lost my phone }\\
    3&   & flow: reset\_2fa; agent: I am so sorry to hear you lost your phone.  Do you need to get your username and password or do you have one of them? & flow: reset\_2fa; agent: \red{I'm sorry to hear that. Let me see what I can do to help.} & agent: Hi! ... customer: yes, i can't access my account because i lost my phone flow: reset\_2fa; action\_plan: pull-up-account, enter-details, send-link;  \\ \hline
    4& \multicolumn{3}{l}{customer: can i give you my phone number instead }\\
     5&   & flow: reset\_2fa; agent: Can I first start with your full name? & flow: reset\_2fa; agent: \red{Sure!} & agent: Hi! ... customer: can i give you my phone number instead flow: reset\_2fa; action\_plan: pull-up-account, enter-details, send-link;  \\ \hline
    6& \multicolumn{3}{l}{customer: Albert Sanders }\\
    7&  &  flow: reset\_2fa; action: pull-up-account: albert sanders & flow: reset\_2fa; action: pull-up-account: albert sanders & agent: Hi! ... customer: Albert Sanders flow: reset\_2fa; action\_plan: pull-up-account, enter-details, send-link;  \\ \hline
    8&   & flow: reset\_2fa; agent: Can you also provide me with your phone number and email address & flow: reset\_2fa; agent: \red{Thank you Albert. I was able to pull up your account.} & agent: Hi! ... action: pull-up-account: albert sanders flow: reset\_2fa; action\_plan: enter-details, send-link;  \\ \hline
    9& \multicolumn{3}{l}{customer: i don't have an email. the phone number is 330-822-4754 }\\
     10&  & flow: reset\_2fa; agent: Ok, that is fine.  What is your zip code & flow: reset\_2fa; \red{action: enter-details: 330-822-4754} & agent: Hi! ... customer: i don't have an email. the phone number is 330-822-4754 flow: reset\_2fa; action\_plan: enter-details, send-link;  \\ \hline
     11&  & flow: reset\_2fa; action: verify-identity: albert sanders, 69233, 330-822-4754 & flow: reset\_2fa; \red{action: enter-details: 69233} & agent: Hi! ... customer: 69233 flow: reset\_2fa; action\_plan: enter-details, send-link;  \\ \hline
      12& & flow: reset\_2fa; agent: Thank you for all the information.  Your username is asanders1 & flow: reset\_2fa; agent: \red{What is your username and password?} & agent: Hi! ... action: verify-identity: albert sanders, 69233, 330-822-4754 flow: reset\_2fa; action\_plan: enter-details, send-link;  \\ \hline
    13&\multicolumn{3}{l}{customer: ha..of course }\\
    14&   & flow: reset\_2fa; agent: I can not actually view your old password for security reason so I have to create a new one & flow: reset\_2fa; \red{action: verify-identity: 69233, asanders1, asanders1} & agent: Hi! T... customer: ha..of course flow: reset\_2fa; action\_plan: enter-details, send-link;  \\ \hline
    15&\multicolumn{3}{l}{customer: ok  }\\
     16&  & flow: reset\_2fa; agent: Can I have your account pin or the answer to your security question & flow: reset\_2fa; \red{agent: enter-details: asanders1} & agent: Hi! ... agent: I can not actually view your old password for security reason so I have to create a new one customer: ok flow: reset\_2fa; action\_plan: enter-details, send-link;  \\ \hline
    17&\multicolumn{3}{l}{customer: security question answer is Alexander }\\        
     18& &  flow: reset\_2fa; agent: Thank you. One moment while I generate the password & flow: reset\_2fa; \red{action: verify-identity: 69233, asanders1, asanders1} & agent: Hi! ...customer: security question answer is Alexander flow: reset\_2fa; action\_plan: enter-details, send-link;  \\ \hline
      19& & flow: reset\_2fa; action: make-password:  & flow: reset\_2fa; action: \red{enter-details: asanders1} & agent: Hi! ... agent: Thank you. One moment while I generate the password flow: reset\_2fa; action\_plan: enter-details, send-link;  \\ \hline
      20& & flow: reset\_2fa; agent: Your password is rox6fnwo33e & flow: reset\_2fa; \red{action: enter-details: asanders1} & agent: Hi! ... action: make-password:  flow: reset\_2fa; action\_plan: enter-details, send-link;  \\ \hline
    21&\multicolumn{3}{l}{customer: ok great}\\ 
    22&   & flow: reset\_2fa; agent: did you have any other questions today? & flow: reset\_2fa; action: \red{enter-details: rox6fnwo33e} & agent: Hi! ... customer: ok great flow: reset\_2fa; action\_plan: enter-details, send-link;  \\ \hline
    23&\multicolumn{3}{l}{customer: nope. thanks so much!} \\
     24&  & flow: reset\_2fa; agent: You're Welcome.  Have a great day & flow: reset\_2fa; agent: \red{Thank you for shopping with Acme. Have a great day} & agent: Hi! ...customer: nope. thanks so much! flow: reset\_2fa; action\_plan: enter-details, send-link; \\ \hline
    \end{tabular}
    \caption{Example dialogue with unseen action make-password. Split-3, D31-1, T5-large, seed: 0, conversation id: 2049. Incorrect predictions are in red.}
    \label{example_output/convo_2049_large}
\end{table*}

\section{ABCD Dataset}

A examples of the ABCD data, and some issues of the dataset.

\subsection{Example workflow instructions}
\label{appendix:abcd_dataset}
An example of the instructions provided to the WoZ for the workflow \code{Initiate Refund} is provided in Listing \ref{example/guidelines_initiate_refund}. Note that no machine-parsable version of these instructions exists in the task labels.

\begin{lstlisting}[style=quote_style, language={}, columns=fullflexible,
label=example/guidelines_initiate_refund,
caption=Sample guidelines provided to the WoZ for the \code{Initiate Refund} workflow.
]
Initiate Refund: 
Start by asking for the refund reason. Then, to complete the refund, perform the following actions: 
	[Interaction] All chats in this flow start by asking the customer for their full name or Account ID using [Pull up Account].  This will pull up their account in the background with the right information related to this user.
	[Kb query] To confirm that their purchase is valid with [Validate Purchase]:
	    Username - occasionally, the customer may not remember their username.  In this case, use the Recover Username subflow in the Account Access flow 
	    Email Address
	    Order ID - last item, to be consistent with verify identity
	[Interaction] Refund method in [Record Reason] - valid options are
	    Gift card - they want a prepaid gift card, also ask for their address so you know where to mail it.  Enter this value in the next step.
	    Add value - to add value to their account.
	    Paper check - also ask for their address so you know where to mail the check.  Enter this address in the next step.
	    Credit card - direct refund to their credit card, assume a credit card is already on file.  To find it, you will need the account ID, which is entered in the next step.
	[Interaction] Add in additional information using [Enter Details]
	    If the customer chose gift card or paper check then enter the full address
	    If the customer chose add value or credit card then enter the account ID
	[Interaction] Dollar amount - enter this value into the details form (without the dollar sign)
	    Then click [Offer Refund]
	    If the customer does not know, the default amount is $50
As always, wrap up by nicely asking if the customer needs any further assistance.
\end{lstlisting}

\subsection{Dataset Issues}
\label{sec:dataset-issues}
The ABCD dataset is not perfect and contains the following issues: 

1) WoZ agents do not always follow the "correct" flow, at least as provided by the labels. This likely arises from miscommunication and realistic gap in knowledge between the WoZ and the customer: the former knows what actions are possible, but not what the customer needs, while the latter knows (hopefully) what they need, but not which actions/flows this requires executing.

2) There are some errors in the transcription of information. For example, the WoZ provides a slot value that contains a typo in the username provided by the customer. The result is that a model that perfectly copies the username from the context will be penalized for not reproducing the error. 

3) Slot-type is not labelled in the targets. For an action like \code{enter-details} for which the agent need only provide 1/7 possible slot values, we are provided with a list with e.g. \code{555-5555} but no indication of whether this single slot was a \code{username}, \code{account\_id}, etc. 

4) Delexicalization is imperfect. If a phone-number is provided in the customer's "scenario", it will be delexicalized in the delexicalized conversation. However, if there is a typo or if the customer makes up the phone-number (as is often the case), then it will not be delexicalized. It is for this reason that we train on the original (not delexicalized) data.

5) The goals and instructions provided to the user are not provided in the dataset, although some user-specific information is provided in the "scenario."

6) The guidelines have extra information about which actions must be taken that are not reflected in the ontology. For example, sometimes, certain slots are required depending on a user's response.

7) Some actions \textit{do} produce effects on conversation state, but these are not captured in the labels. For example, "ask-the-oracle" returns a yes/no value to the Woz, which determines the action they should take and actually alters the workflow, but that returned value can at best be inferred through the rest of the conversation.

\section{Flow prediction accuracy across turns}
\label{sec:flow-prediction-accuracy-across-turns}
In practice, the high level intent or goal of a user may not be apparent until multiple turns of the dialogue have passed.  We hypothesize that flow prediction accuracy will be limited early in the dialogue, but will improve as the conversation progresses and more information is uncovered about the user and their issue. Similarly, previously performed actions could provide clues to the overall flow, although incorrect actions could also provide incorrect signals.

\begin{figure}[h]
    \centering
    \subfloat[Flow accuracy]{
    \includegraphics[clip,width=\columnwidth]{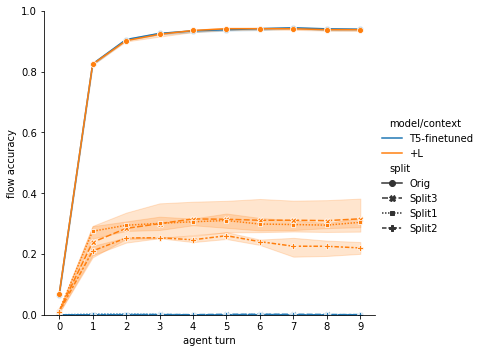}
    }
    
    \subfloat[Flow prefix accuracy]{
    \includegraphics[clip,width=\columnwidth]{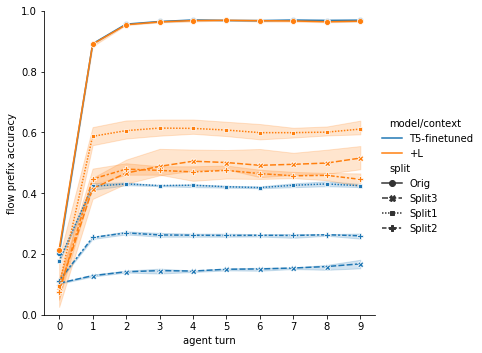}
    }
    \caption{Flow and Flow prefix accuracy for T5-large finetuned (blue) and T5-large finetuned + L (orange) (with legal flow list provided) models across agent turns (x-axis) and across splits (line style and marker style).}
    \label{fig:flow_prefix_accuracy_per_turn}
\end{figure}

Figure \ref{fig:flow_prefix_accuracy_per_turn} shows that it is indeed the case that flow prediction accuracy improves over the first few turns of the conversation. On the standard train/test split, whether or not the legal flow list is provided, flow prediction accuracy increases from <10\% on turn 0 to plateau at \quasi90\% %
by turn 3. %

For the other dataset splits, in the zero-shot flow prediction scenario, the same basic trend is apparent for the model provided with the legal list of flows (+L): performance increases to \quasi20-30\% %
by turn \quasi2. Without the list of flows (+$\emptyset$, shown in blue), flow prediction remains \quasi0\% across turns. 

\textit{Flow prefix} prediction shows a similar trend, with accuracy plateauing at turn \quasi1-2 and improving by providing the flow list (+L). Per turn prefix accuracy decreases from Split 1 to 2 and 3, since there are fewer similarities between the flow prefixes at train and test time.

The above suggests that it will be important in practice to allow for flow predictions to be ignored (or not made) early in the conversation, and to not commit to plans until sufficient information exists. %

See similar graphs for T5-base in Appendix~\ref{t5_small_results}.

\section{Slot prediction}
\label{sec:slot_metrics}

Here we study slot accuracy metrics as
models are trained to output slot values. 
As with flow information, slot types could be provided in the context as part of an ontology. Here we show a preliminary explorations along these lines.

\subsection{Evaluation}
The metrics for slot prediction are provided in Table~\ref{tab:large-slot-result-overall} and are as follows: 
\begin{itemize}[noitemsep,topsep=0pt]
    \item \textbf{Slot (mean/all)}: slot value accuracy, average of per-turn slot correctness for ground-truth action-turns only. Per-turn, slots are considered correct if they are provided in the \textit{exact same order as in the ground truth}. Per-turn slot correctness is either the average number of correct slots per turn (\textit{mean}) or 1/0 if all slots are/are not correct (\textit{all}).  
\end{itemize}

\subsection{Results}

\begin{table}[]
    \footnotesize
    \centering
    \resizebox{0.48\textwidth}{!}{
    \centering
    \begin{tabular}{l c}
        \hline
        {Model} + Additional Context & {Slot mean/all} \\
        \hline
         \multicolumn{2}{l}{\textbf{Standard split: all actions and workflows seen in training}} \\
        \hline
        T5-large finetuned &89.0 / 88.3  \\
        + L &89.3 / 88.6 \\
        \arrayrulecolor{lightgray}\hline
        + F &89.2 / 88.3 \\
        + L + F &89.5 / 88.7 \\
        \arrayrulecolor{lightgray}\hline    
        + F+P &89.4 / 88.6 \\
        + L + F+P &89.6 / 88.7 \\
        \hline
        \multicolumn{2}{l}{\textbf{Split 1: All flows are unseen in training}} \\ 
        \hline
        T5-large baseline &84.6 / 84.8 \\
        + L &84.6 / 84.9 \\
        \arrayrulecolor{lightgray}\hline 
        + F &85.3 / 85.5 \\
        + L + F &85.0 / 85.4 \\
        \arrayrulecolor{lightgray}\hline 
        + F + P &87.6 / 87.1 \\
        + L + F + P &86.5 / 86.2 \\
        \arrayrulecolor{lightgray}\hline 
        
        \hline
         \multicolumn{2}{l}{\textbf{Split 2: Flow prefix groups and action sequences unseen in training}} \\
        \hline
        T5-large baseline &81.6 / 81.6 \\
        + L  &82.5 / 82.7 \\
        \arrayrulecolor{lightgray}\hline 
        + F &81.9 / 82.0 \\
        + L + F &82.9 / 82.8 \\
        \arrayrulecolor{lightgray}\hline 
        + F + P  &83.0 / 82.6 \\
        + L + F + P &83.7 / 83.2 \\
        \hline
        
         \multicolumn{2}{l}{\textbf{Split 3: Flow prefix groups and action sequences unseen in training}} \\
        \hline
        T5-large baseline &74.7 / 77.8 \\
        + L &74.7 / 77.9 \\
         \hline
        + F &73.4 / 76.6 \\
        + L + F &72.8 / 75.5 \\
        \arrayrulecolor{lightgray}\hline 
        + F + P &75.1 / 77.1 \\
        + L + F + P &73.9 / 76.2 \\ 
        \hline
    \end{tabular}
    }
    \caption{Slot prediction metrics for models trained across various generalization splits explained in
Section \ref{sec:eval_method}. The Baseline models are finetuned T5-large models and model variants provide different amounts of additional context beyond
the dialogue utterances described in Section \ref{sec:augmenting_context}, consisting of: legal flow lists (L), flow (last turn) (F), or the action sequence plan
(last turn) (P)}
    \label{tab:large-slot-result-overall}
\end{table}

As with action prediction, slot prediction accuracy differs little across the different context augmentations for the standard train/test split ("slot mean" at \quasi89.5\% for T5-large). 
The "slot all" metric differs little from "slot mean", suggesting that there is infrequent confusion in placement of slot values, and is likely since most actions require only one slot. 

Unlike action prediction, we do not see a large difference in slot accuracy across the different settings for the various generalization splits. %
However, the overall decrease in slot accuracy across splits may suggest that it is challenging to determine the correct slot values for unknown actions or actions in unknown flows. An example of this is provided in Example \ref{example_output/convo_6601}. Altogether, this suggests that different kinds of information (e.g. slot types) would be required to improve slot-prediction accuracy.

We ran preliminary experiments with more generous slot-prediction metrics and with providing slot type information along with the action plan but no improvement in slot performance was observed (see Appendices~\ref{sec:slot_types} and ~\ref{sec:generous_slot_metrics}).

\section{T5-small and T5-base Results}
\label{t5_small_results}

The following results for T5-small and T5-base combine all previously discussed metrics, including the Slot prediction metrics presented in Appendix~\ref{sec:slot_metrics}.
The results for our T5-large based models in Table \ref{tab:large-result-overall}.

\begin{table}[h]
    \centering
    \resizebox{0.5\textwidth}{!}{
    \begin{tabular}{lccc}
        \hline
        Model configuration & \% train & \% test only\ & \% neither  \\
        \hline
        T5-base finetuned & 95.6 & 0.01 & 4.3 \\
        + L & 71.9 & 22.8 & 5.4 \\
        \hline
    \end{tabular}
    }
    \caption{Source of predicted flows for T5-base models on Split 3 (averaged over 3 seeds)}
    \label{tab:flow_from_where_base}
\end{table}

\begin{table}[h]
    \footnotesize
    \centering
    \resizebox{0.5\textwidth}{!}{
    \centering
    \begin{tabular}{l p{0.8cm} c p{0.8cm} p{0.8cm} p{1cm}}
        \hline
        Model & Action & Slot mean/all & Flow & Flow prefix & Lev action/Lev action free del \\
        \hline
         \multicolumn{6}{l}{Standard: all actions and workflows seen in training} \\
        \hline
        T5-small finetuned & 76.7 & 77.3 / 75.5 & 84.1 & 89.0 & 4.4 / 0.5\\
        + L & 76.5 & 78.5 / 76.3 & 83.6 & 88.6 & 4.4 / 0.5\\
        + F & 77.4 & 79.0 / 77.2 & 93.5 & 94.5 & 3.9 / 0.5\\
        + L + F & 75.3 & 78.8 / 77.5 & 93.5 & 94.4 & 3.8 / 0.5\\
        + F+P & 77.3 & 78.2 / 76.4 & 93.5 & 94.5 & 4.1 / 0.5\\
        + L + F+P & 77.7 & 78.5 / 76.6 & 93.5 & 94.2 & 4.4 / 0.5\\
        \hline
         \multicolumn{6}{l}{Split 1: All flows are unseen in training} \\
        \hline
        T5-small finetuned & 64.9 & 74.6 / 73.4 & 2.8 & 47.2 & 4.7 / 0.9\\
        + L & 62.6 & 74.5 / 73.6 & 19.2 & 51.0 & 5.6 / 0.9\\
        + F & 63.0 & 75.2 / 74.0 & 93.4 & 94.6 & 4.2 / 0.9\\
        + L + F & 59.9 & 74.8 / 74.1 & 93.4 & 94.4 & 4.2 / 1.1\\
        + F+P & 73.2 & 77.1 / 75.4 & 93.4 & 94.6 & 4.4 / 0.6\\
        + L + F+P & 71.2 & 78.6 / 77.0 & 93.4 & 94.6 & 4.1 / 0.7\\
        \hline
         \multicolumn{6}{l}{Split 2: Flow prefix groups and action sequences unseen in training} \\
        \hline
        T5-small finetuned  & 58.4 & 71.9 / 71.5 & 0.0 & 26.2 & 5.4 / 1.2\\
        + L  & 53.2 & 70.6 / 72.1 & 5.7 & 32.0 & 5.8 / 1.3\\
        + F  & 55.6 & 72.6 / 71.9 & 93.6 & 94.4 & 4.7 / 1.3\\
        + L + F  & 53.7 & 71.3 / 72.3 & 91.4 & 94.3 & 5.1 / 1.3\\
        + F+P  & 68.8 & 71.4 / 70.2 & 93.5 & 94.4 & 5.1 / 0.8\\
        + L + F+P  & 68.0 & 74.0 / 73.3 & 92.3 & 93.9 & 5.0 / 0.9\\
        \hline
         \multicolumn{6}{l}{Split 3: Flow prefix groups and action sequences unseen in training} \\
        \hline
        T5-small finetuned & 44.1 & 66.7 / 69.1 & 0.0 & 16.3 & 5.7 / 1.4\\
        + L & 41.1 & 65.5 / 69.0 & 4.7 & 20.6 & 6.8 / 1.3\\
        + F & 44.1 & 65.7 / 68.2 & 93.1 & 94.0 & 4.5 / 1.5\\
        + L + F & 41.8 & 66.6 / 69.1 & 93.1 & 93.4 & 4.8 / 1.5\\
        + F+P & 64.4 & 65.9 / 66.9 & 93.2 & 94.0 & 4.7 / 0.8\\
        + L + F+P & 64.2 & 64.8 / 65.3 & 91.7 & 93.6 & 5.0 / 0.8\\
        \hline
    \end{tabular}
    }
    \caption{Action, slot, and flow prediction metrics for models trained across various generalization splits explained in
Section 6.1. Models are all finetuned T5-small models provided with different amounts of additional context described in Section 4.4: legal flow list (L), flow (last turn) (F), action sequence plan
(last turn) (P)}
    \label{tab:small-result-overall}
\end{table}

\begin{table}[h]
    \footnotesize
    \centering
    \resizebox{0.5\textwidth}{!}{
    \centering
    \begin{tabular}{l p{3cm} p{3cm}}
        \hline
        Model & Theoretical unseen / seen actions & Actual unseen / seen actions \\
        \hline
        T5-small finetuned & 1.3 / 53.6 & 0.0 / 45.0 \\
        + L & 5.7 / 48.9 & 0.0 / 41.9 \\
        + F & 17.9 / 50.0 & 0.0 / 45.0 \\
        + L + F & 8.0 / 49.2 & 0.0 / 42.6 \\
        + F + P & 55.9 / 66.3 & 74.6 / 64.2 \\
        + L + F + P & 56.3 / 65.9 & 75.4 / 64.0 \\
        \hline
        \end{tabular}
        }
    \caption{Results - Split 3: T5-small Flow prefix groups and action sequences unseen in training}
    \label{tab:small-result-split-3-seen-vs-unseen}
\end{table}

\begin{table}[h]
    \small
    \centering
    \resizebox{0.5\textwidth}{!}{
    \centering
    \begin{tabular}{p{2.3cm} p{0.8cm} c p{0.8cm} p{0.8cm}  p{1cm}}
        \hline
        Model + Additional Context & Action & Slot mean/all & Flow & Flow prefix & Lev action/Lev action free del \\
        \hline
        \multicolumn{6}{l}{\rule{0pt}{10pt}\textbf{Standard:} all actions and workflows seen in training } \\
        \hline
        T5-base finetuned & 83.2 & 86.1 / 85.3 & 85.2 & 89.7 & 3.2 / 0.4\\
        + L & 83.8 & 86.4 / 85.6 & 85.2 & 89.6 & 3.3 / 0.3\\
        \arrayrulecolor{lightgray}\hline
        + F & 82.9 & 86.5 / 85.7 & 93.5 & 94.5 & 3.2 / 0.4\\ 
        + L + F & 83.1 & 85.7 / 85.0 & 93.5 & 94.4 & 3.2 / 0.4\\
        \arrayrulecolor{lightgray}\hline
        + F + P & 83.0 & 85.7 / 84.8 & 93.5 & 94.5 & 3.3 / 0.4\\
        + L + F + P & 83.5 & 85.8 / 84.9 & 93.5 & 94.4 & 3.3 / 0.4\\
        \arrayrulecolor{lightgray}\hline
        \arrayrulecolor{black}\hline
        \multicolumn{6}{l}{\rule{0pt}{10pt}\textbf{Split 1:} All flows are unseen in training} \\
        \hline
        T5-base finetuned & 72.4 & 82.6 / 82.9 & 0.1 & 42.6 & 3.4 / 0.8\\
        \arrayrulecolor{lightgray}\hline
        + L & 73.6 & 82.9 / 83.1 & 25.6 & 54.0 & 3.5 / 0.7\\
        + F & 72.6 & 82.2 / 82.3 & 92.5 & 94.6 & 3.5 / 0.8\\
        \arrayrulecolor{lightgray}\hline
        + L + F & 73.4 & 82.1 / 82.4 & 93.4 & 94.6 & 3.5 / 0.7\\
        + F + P & 79.8 & 85.8 / 85.3 & 92.8 & 94.2 & 3.4 / 0.5\\
        \arrayrulecolor{lightgray}\hline
        + L + F + P & 81.0 & 83.9 / 83.4 & 93.4 & 94.6 & 3.6 / 0.5\\
        \arrayrulecolor{black} \hline
        \multicolumn{6}{l}{\rule{0pt}{10pt}\textbf{Split 2:} Flow prefix groups and action sequences unseen in training} \\
        \hline
        T5-base finetuned & 67.6 & 78.9 / 79.4 & 0.1 & 24.3 & 4.1 / 1.0\\
        \arrayrulecolor{lightgray}\hline
        + L  & 68.6 & 78.4 / 79.0 & 14.4 & 36.9 & 3.9 / 1.0\\
        + F  & 69.2 & 78.5 / 78.7 & 93.1 & 94.4 & 4.0 / 0.9\\
        \arrayrulecolor{lightgray}\hline
        + L + F  & 68.7 & 78.3 / 78.5 & 93.7 & 95.0 & 4.1 / 0.9\\
        + F + P  & 77.6 & 77.6 / 77.2 & 93.3 & 94.4 & 4.0 / 0.6\\
        \arrayrulecolor{lightgray}\hline
        + L + F + P  & 76.9 & 78.1 / 77.7 & 93.7 & 94.6 & 3.9 / 0.6\\
        \arrayrulecolor{black}\hline
        \multicolumn{6}{l}{\rule{0pt}{10pt}\textbf{Split 3:} Flow prefix groups and action sequences unseen in training} \\
        \hline
        T-base finetuned & 52.6 & 70.9 / 75.1 & 0.1 & 15.7 & 4.3 / 1.3\\
        \arrayrulecolor{lightgray}\hline
        + L & 53.1 & 70.6 / 74.8 & 17.8 & 35.6 & 4.0 / 1.3\\
        + F & 54.3 & 69.6 / 73.8 & 93.2 & 94.0 & 3.8 / 1.3\\
        \arrayrulecolor{lightgray}\hline
        + L + F & 54.5 & 69.8 / 73.6 & 93.2 & 93.2 & 3.9 / 1.2\\
        + F + P & 65.4 & 72.6 / 75.3 & 93.2 & 94.0 & 3.5 / 0.8\\
        \arrayrulecolor{lightgray}\hline
        + L + F + P & 66.8 & 72.1 / 74.8 & 93.2 & 93.3 & 3.7 / 0.8\\
    \end{tabular}
    }
    \caption{Action, slot, and flow prediction metrics for models trained across various generalization splits explained in Section \ref{sec:eval_method}. Models are all finetuned T5-base models provided with different amounts of additional context described in Section \ref{sec:augmenting_context}: legal flow list (L), flow (last turn) (F), action sequence plan (last turn) (P).}
    \label{tab:base-result-overall}
\end{table}

\begin{table}[h]
    \footnotesize
    \centering
    \resizebox{0.5\textwidth}{!}{
    \centering
    \begin{tabular}{p{2.5cm} p{1.5cm} p{1.5cm} p{1.5cm} p{1.5cm}}
        \hline
        Model + Additional Context  & Theoretical seen  & Theoretical unseen  & Actual seen & Actual unseen \\
        \hline
        T5-base finetuned  & 59.5 &  21.5 & 53.7 & 0.0  \\
        + L  & 59.8 & 23.1  & 54.2 & 0.0  \\
        + F  & 59.2 & 32.0  & 55.4 & 0.0  \\
        + L + F  & 59.6 & 31.7 &55.6 &  0.0 \\
        + F+P  & 68.2 & 52.6 & 65.1 & 79.4 \\
        + L + F+P  & 69.5 & 54.4 & 66.5 & 80.2  \\
        \hline
        \end{tabular}
        }
        
    \caption{Results - Split 3: Flow prefix groups and action sequences unseen in training for T5-base}
    \label{tab:result-split-3-seen-vs-unseen}
\end{table}

\begin{figure}[h]
    \centering
    \includegraphics[width=0.5\textwidth]{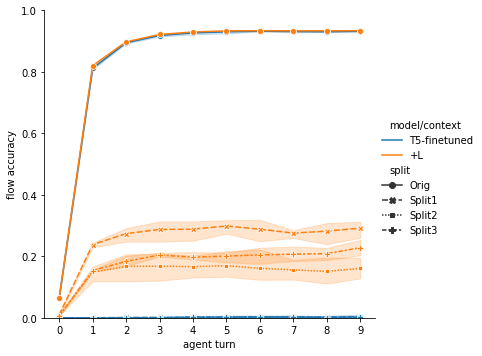}
    \caption{Flow accuracy for T5-base finetuned (blue) and T5-base finetuned + L (orange) (with legal flow list provided) models across agent turns (x-axis) and across splits (line style and marker style).}
    \label{fig:flow_accuracy_per_turn_base}
\end{figure}

\begin{figure}[h]
    \centering
    \includegraphics[width=0.5\textwidth]{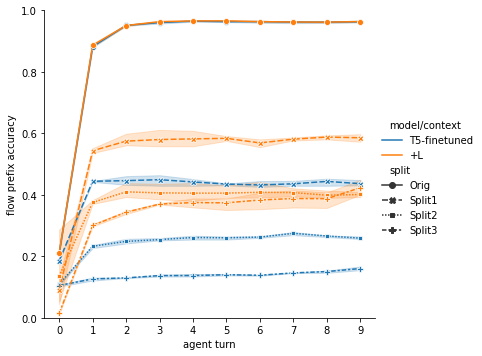}
    \caption{Flow prefix accuracy for T5-base finetuned (blue) and T5-base finetuned + L (orange) (with legal flow list provided) models across agent turns (x-axis) and across splits (line style and marker style).}
    \label{fig:flow_prefix_accuracy_per_turn_base}
\end{figure}

\section{Select experiments with slot types and more generous slot-metric prediction}

We ran preliminary experiments with more generous slot-prediction metrics and with providing slot type information along with the action plan but no improvement in slot performance was observed. They are in section~\ref{sec:slot_types} and section ~\ref{sec:generous_slot_metrics}).

\subsection{Select experiments with providing slot types}
\label{sec:slot_types}

\begin{table}[h]
    \small
    \centering
    \resizebox{0.5\textwidth}{!}{
    \centering
    \begin{tabular}{p{2.3cm} p{0.8cm} c p{0.8cm} p{0.8cm}  p{1cm}}
        \hline
        Model + Additional Context & Action & Slot mean/all & Flow & Flow prefix & Lev action/with free del \\
        \hline
        \multicolumn{6}{l}{\rule{0pt}{10pt}\textbf{Standard:} all actions and workflows seen in training } \\
        \hline
        + F + P() &  83.4 & 85.7 / 84.7 & 93.5 & 94.5 & 3.3 / 0.4 \\
        + L + F + P() & 83.6 & 86.0 / 85.1 & 93.5 & 94.4 & 3.2 / 0.4 \\
        \arrayrulecolor{black}\hline
        \multicolumn{6}{l}{\rule{0pt}{10pt}\textbf{Split 1:} All flows are unseen in training} \\
        \hline
        + F + P() & 80.4 & 85.0 / 84.8 & 91.0 & 94.6 & 3.4 / 0.5 \\
        + L + F + P() & 81.8 & 84.3 / 83.9 & 93.3 & 94.6 & 3.4 / 0.5 \\
        \arrayrulecolor{black} \hline
        \multicolumn{6}{l}{\rule{0pt}{10pt}\textbf{Split 2:} Flow prefix groups and action sequences unseen in training} \\
        \hline
        + F + P() & 77.3 & 77.4 / 77.2 & 92.1 & 94.2 & 3.8 / 0.6 \\
        + L + F + P() & 77.4 & 78.6 / 78.3 & 93.2 & 94.0 & 3.8 / 0.6 \\
        \arrayrulecolor{black}\hline
        \multicolumn{6}{l}{\rule{0pt}{10pt}\textbf{Split 3:} Flow prefix groups and action sequences unseen in training} \\
        \hline
        + F + P() & 68.2 & 71.7 / 74.3 & 92.9 & 94.0 & 4.2 / 0.8 \\
        + L + F + P() & 68.1 & 73.6 / 76.4 & 93.2 & 93.4 & 4.0 / 0.8 \\        
        \hline
    \end{tabular}
    }
    \caption{Action, slot, and flow prediction metrics for models trained across various generalization splits explained in Section \ref{sec:eval_method}. Models are all finetuned T5-base models provided with different amounts of additional context described in Section \ref{sec:augmenting_context}: legal flow list (L), flow (last turn) (F), action sequence plan (last turn) \textit{with slot types} (P()).}
    \label{tab:base-result-overall-adding-slottypes}
\end{table}

\begin{table}[h!]
    \footnotesize
    \centering
    \resizebox{0.5\textwidth}{!}{
    \centering
    \begin{tabular}{p{2.3cm} p{0.8cm} c p{0.8cm} p{0.8cm} p{1cm}}
        \hline
        Model + Additional Context & Action & Slot mean/all & Flow & Flow prefix & Lev action/Lev action free del \\
        \hline
         \multicolumn{6}{l}{\textbf{Standard:} all actions and workflows seen in training} \\
        \hline
        + F + P() & 85.2 & 89.5 / 88.7 & 93.5 & 94.5 & 2.8 / 0.3 \\
        + L + F + P() & 85.4 & 89.6 / 88.8 & 93.5 & 94.5 & 2.8 / 0.3 \\
        \hline
        \multicolumn{6}{l}{\textbf{Split 1}: All flows are unseen in training} \\ 
        \hline
        + F + P() &  82.3 & 86.5 / 86.3 & 93.3 & 94.6 & 2.8 / 0.4 \\
        + L + F + P() & 82.9 & 87.7 / 87.6 & 92.9 & 94.5 & 2.8 / 0.4 \\
        
        \hline
         \multicolumn{6}{l}{\textbf{Split 2}: Flow prefix groups and action sequences unseen in training} \\
        \hline
        + F + P() & 80.4 & 81.8 / 81.7 & 93.2 & 94.4 & 3.2 / 0.5 \\
        + L + F + P() & 80.9 & 82.5 / 82.2 & 92.5 & 94.4 & 3.1 / 0.5 \\         
        \hline
        
         \multicolumn{6}{l}{\textbf{Split 3}: Flow prefix groups and action sequences unseen in training} \\
        \hline
         + F + P() & 70.7 & 73.9 / 75.7 & 90.9 & 93.9 & 3.4 / 0.7 \\
         + L + F + P() & 70.8 & 73.0 / 75.2 & 93.2 & 93.7 & 3.5 / 0.7 \\       
        \arrayrulecolor{black}\hline
    \end{tabular}
    }
    \caption{Action, slot, and flow prediction metrics for models trained across various generalization splits explained in
Section \ref{sec:eval_method}. Models are all finetuned T5-large models provided with different amounts of additional context described in Section \ref{sec:augmenting_context}: legal flow list (FL), flow (last turn) (F), action sequence plan
(last turn) \textit{with slot types} (P())}
    \label{tab:large-result-overall-with-slot-types}
\end{table}

\subsection{Experiments with generous slot metrics}
\label{sec:generous_slot_metrics}
The slot accuracy metrics described previously require that slot values are provided in the exact same order as provided in the ground truth. However, this is not strictly required for correctness and there is no restriction on the order input by the WoZ in the ground-truth. Thus, below, we provide more generous slot-metrics. These measure accuracy of the slot values \textit{regardless of order}: 

\begin{itemize}
    \setlength\itemsep{0em}
    \item \textbf{Slot set (denom=num expected)}: number of correct slot values divided by the total number of expected slot values in the ground truth. This can be calculated by including or excluding actions with 0 slots.
    \item \textbf{Slot set (denom=num predicted)}: number of correct slot values divided by the total number of predicted slots values
    \item \textbf{Slot set (denom=num longest)}: number of correct slot values divided by the greater of the number of predicted slots and expected slots. This can be calculated by including or excluding actions with 0 slots.
\end{itemize}

\begin{table}[h!]
    \footnotesize
    \centering
    \resizebox{0.5\textwidth}{!}{
    \centering
    \begin{tabular}{p{3cm} p{1.8cm} p{1.8cm} p{1.8cm} }
    \arrayrulecolor{black}\hline
Model +Additional Context & Slot set (denom=num. expected) / with empty slots & Slot set (denom=num. predicted) & Slot set accuracy (denom=num. longest) / with empty slots \\

\hline
        \multicolumn{4}{l}{\rule{0pt}{10pt}\textbf{Standard:} all actions and workflows seen in training } \\
\hline
+ F + P &  79.7 / 79.7 & 86.7 & 79.5 / 77.0 \\
+ L + F + P & 80.3 / 80.3 & 86.8 & 80.1 / 77.4 \\
+ F + P() &  80.0 / 80.0 & 86.7 & 79.8 / 77.2 \\
+ L + F + P()&  80.3 / 80.3 & 87.1 & 80.1 / 77.6 \\

\hline
        \multicolumn{4}{l}{\textbf{Split 1:} All flows are unseen in training} \\ 
        \hline
+ F + P &  77.8 / 77.8 & 86.9 & 77.7 / 76.0  \\
+ L + F + P &  77.2 / 77.2 & 85.0 & 77.0 / 75.4  \\

+ F + P() &  77.9 / 77.9 & 86.0 & 77.6 / 75.8 \\
+ L + F + P()&  78.0 / 78.0 & 85.1 & 77.8 / 76.1  \\

\hline
         \multicolumn{4}{l}{\textbf{Split 2:} Flow prefix groups and action sequences unseen in training} \\
         \hline
+ F + P &  71.8 / 71.8 & 78.7 & 71.2 / 69.4  \\
+ L + F + P & 71.7 / 71.7 & 79.3 & 71.0 / 69.3  \\

+ F + P() & 71.4 / 71.4 & 78.4 & 70.9 / 68.7  \\
+ L + F + P()&  72.5 / 72.5 & 79.7 & 71.9 / 69.9  \\

\hline
\multicolumn{4}{l}{\textbf{Split 3:} Flow prefix groups and action sequences unseen in training} \\
\hline
+ F + P &  61.1 / 61.1 & 75.0 & 60.3 / 56.2  \\
+ L + F + P &  61.2 / 61.2 & 74.4 & 60.5 / 56.6  \\

+ F + P() &  60.5 / 60.5 & 74.2 & 59.8 / 56.6 \\
+ L + F + P()&  61.0 / 61.0 & 75.9 & 60.3 / 57.5  \\

        \hline
        \end{tabular}
        }
    \caption{Generous set-based slot metrics for model trained across generalization splits explained in Section 6. Models are finetuned T5-base
     provided with different amounts of additional
    context described
    in Section 4.4: legal flow list (L), flow (last turn) (F),
    action sequence plan (last turn) (P), and plan \textit{with slot types} (P())}
    \label{tab:temp}
\end{table}

\begin{table}[]
    \footnotesize
    \centering
    \resizebox{0.48\textwidth}{!}{
    \centering
    \begin{tabular}{p{3cm} p{1.8cm} p{1.8cm} p{1.8cm} }
    \hline
    Model +Additional Context & Slot set (denom=num. expected) / with empty slots & Slot set (denom=num. predicted) & Slot set accuracy (denom=num. longest) / with empty slots \\
    \hline
        \multicolumn{4}{l}{\rule{0pt}{10pt}\textbf{Standard:} all actions and workflows seen in training } \\
        \hline
        + F + P()   & 84.3 / 84.3 & 90.5 & 84.0 / 81.7 \\
        + L + F + P()  & 84.5 / 84.5 & 90.7 & 84.3 / 82.1 \\
        \hline
                \multicolumn{4}{l}{\textbf{Split 1:} All flows are unseen in training} \\ 
                \hline
        + F + P() &  80.4 / 80.4 & 87.5 & 80.2 / 78.3  \\
        + L + F + P()  & 81.0 / 81.0 & 88.6 & 80.8 / 79.0  \\
        \hline
         \multicolumn{4}{l}{\textbf{Split 2:} Flow prefix groups and action sequences unseen in training} \\
        \hline
        + F + P() & 75.8 / 75.8 & 82.8 & 75.4 / 73.4 \\
        + L + F + P()  & 76.3 / 76.3 & 83.5 & 75.9 / 73.9  \\
        \hline
        \multicolumn{4}{l}{\textbf{Split 3:} Flow prefix groups and action sequences unseen in training} \\
        \hline
        + F + P() & 64.0 / 64.0 & 75.9 & 63.3 / 60.3 \\
        + L + F + P()  & 63.5 / 63.5 & 75.2 & 62.9 / 60.0  \\

        \hline
        \end{tabular}
        }
    \caption{Generous set-based slot metrics for model trained across generalization splits explained in Section 6. Models are finetuned T5-large
     provided with different amounts of additional
    context described
    in Section 4.4: legal flow list (L), flow (last turn) (F),
    action sequence plan (last turn) (P), and plan \textit{with slot types} (P())}
    \label{tab:slot metrics}
\end{table}

\section{Action prediction confusion matrices}
\label{sec:action-prediction-confusion-matrices}

\begin{figure*}[!h]
    \centering
    \includegraphics[width=0.98\textwidth]{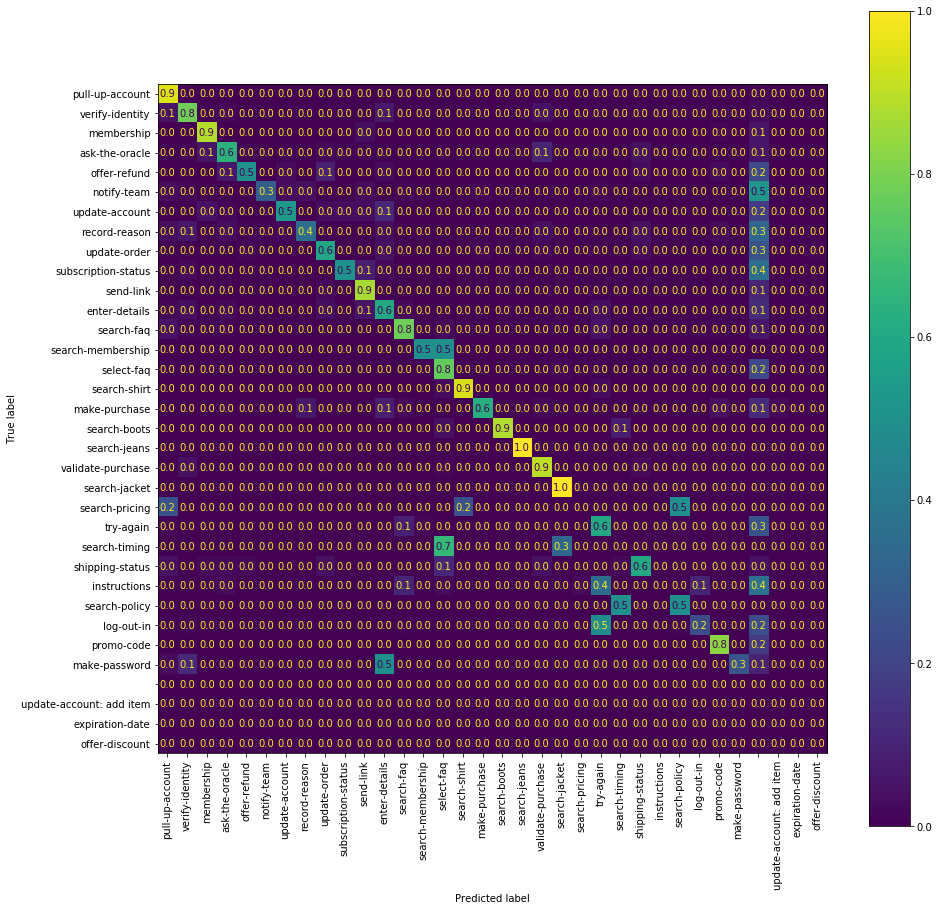}
    \caption{Action confusion matrix for +F+P T5-large finetuned model, seed:0, Split 3}
    \label{fig:action_conf_matrix_large_d31-1_seed0_split3}
\end{figure*}

\begin{figure*}
    \centering
    \includegraphics[width=\textwidth]{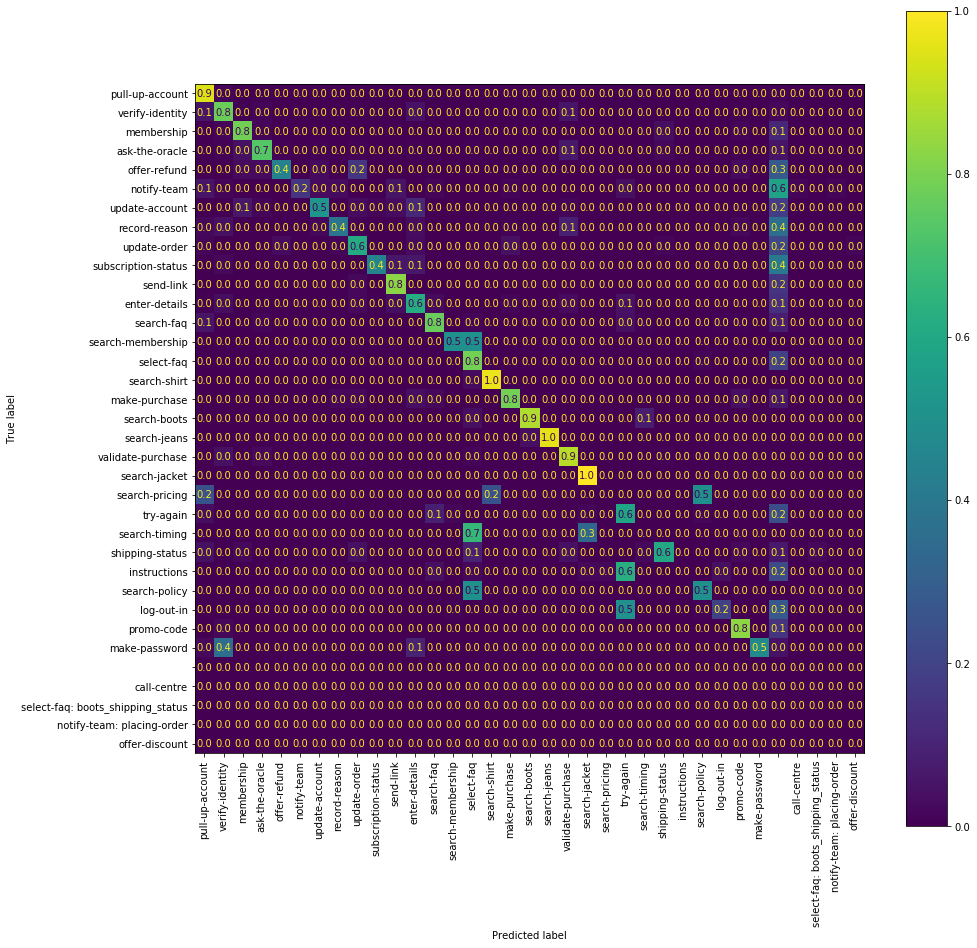}
    \caption{Action confusion matrix for +F+P T5-large finetuned model, seed:0, Split 3}
    \label{fig:action_conf_matrix_large_d31-1_seed1_split3}
\end{figure*}

\begin{figure*}
    \centering
    \includegraphics[width=\textwidth]{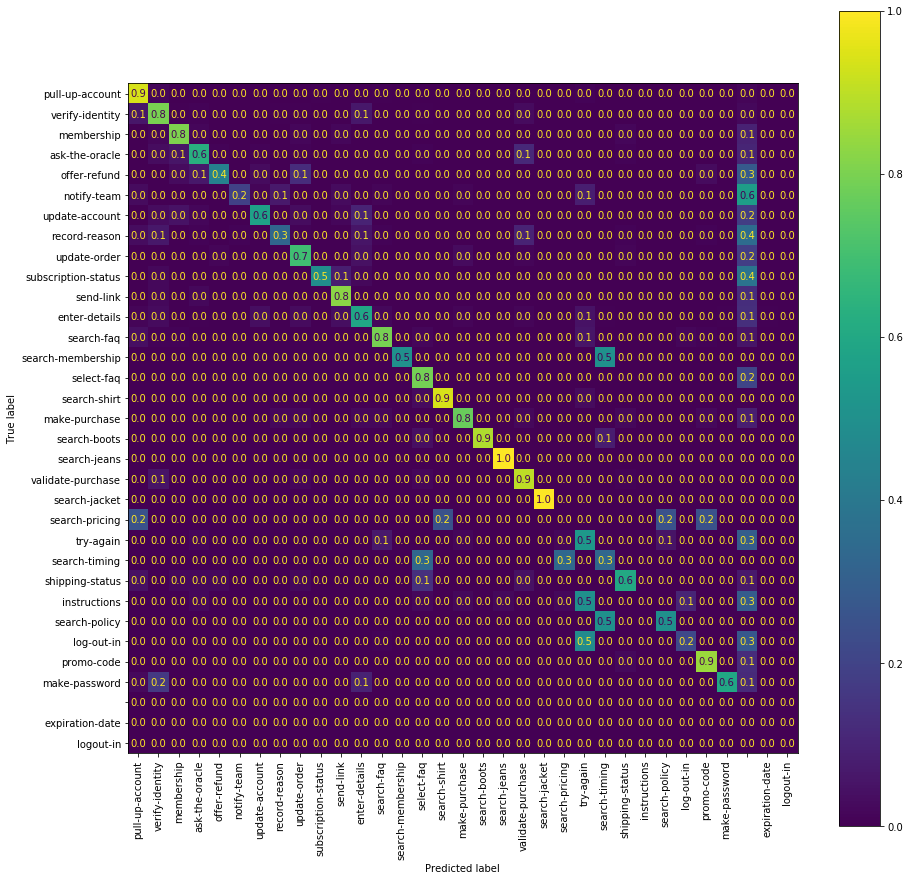}
    \caption{Action confusion matrix for +F+P T5-large finetuned model, seed:42, Split 3}
    \label{fig:action_conf_matrix_large_d31-1_seed2_split3}
\end{figure*}

\begin{figure*}
    \centering
    \includegraphics[width=\textwidth]{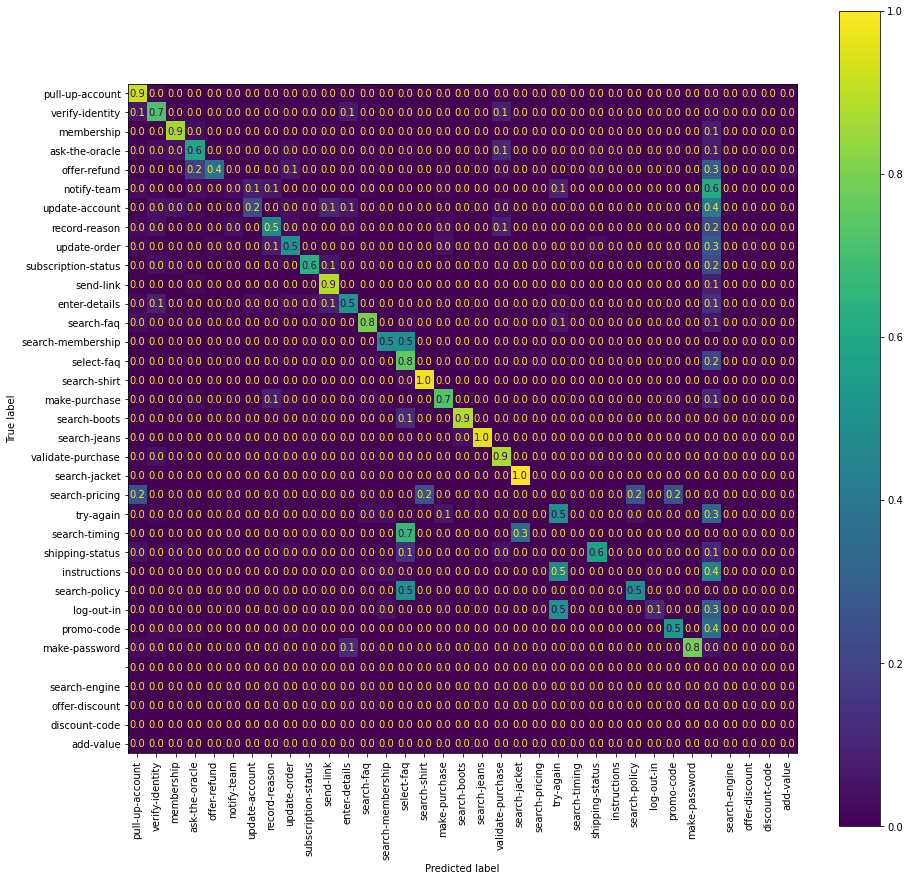}
    \caption{Action confusion matrix for +F+P T5-base finetuned model, seed:0, Split 3}
    \label{fig:action_conf_matrix_d31-1_seed0_split3}
\end{figure*}

\begin{figure*}
    \centering
    \includegraphics[width=\textwidth]{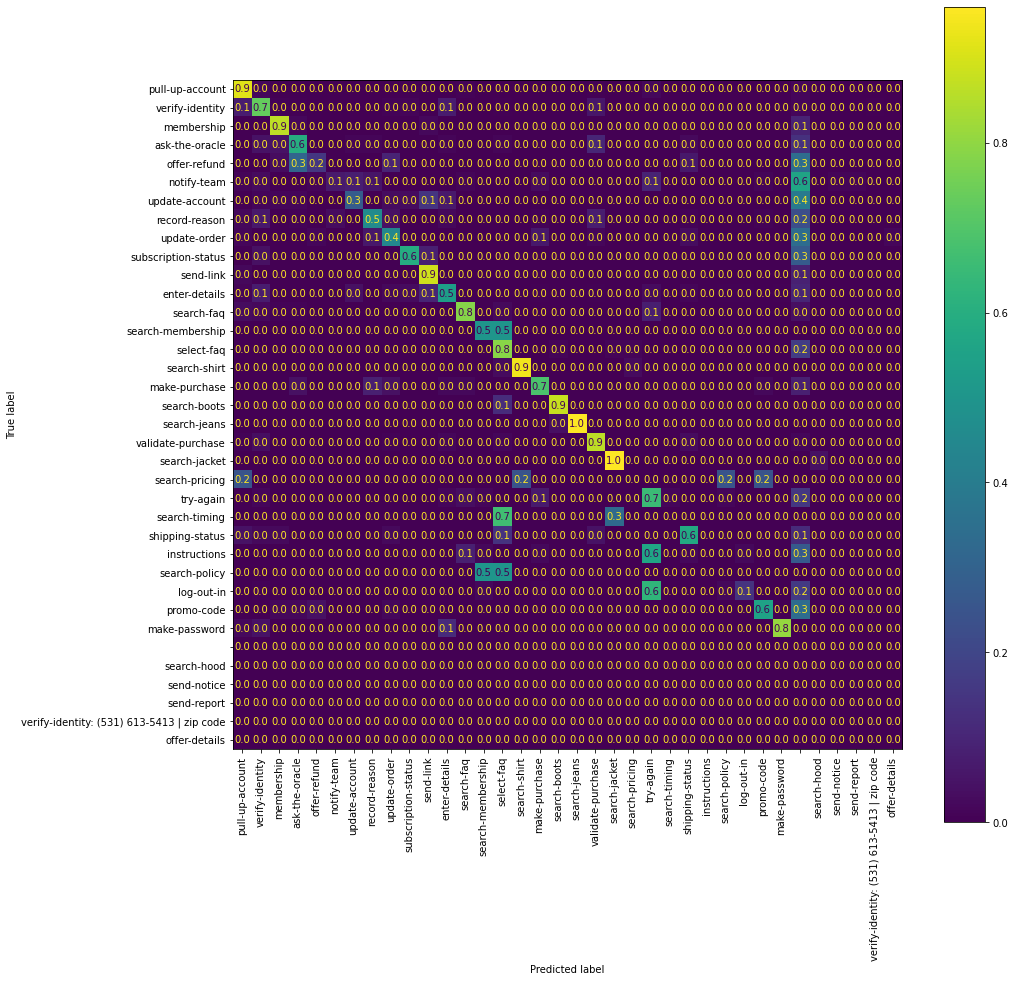}
    \caption{Action confusion matrix for +F+P T5-base finetuned model, seed:0, Split 3}
    \label{fig:action_conf_matrix_d31-1_seed1_split3}
\end{figure*}

\begin{figure*}
    \centering
    \includegraphics[width=\textwidth]{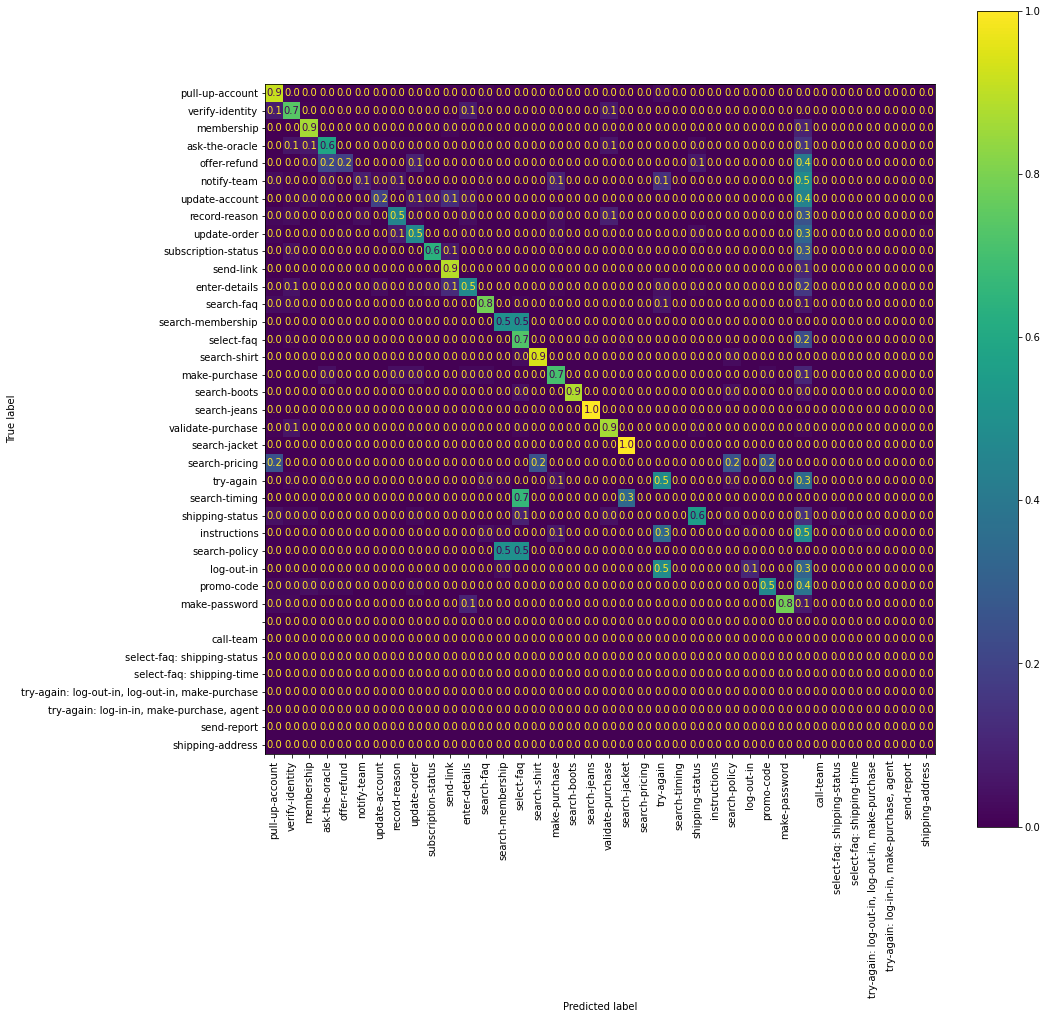}
    \caption{Action confusion matrix for +F+P T5-base finetuned model, seed:42, Split 3}
    \label{fig:action_conf_matrix_d31-1_seed2_split3}
\end{figure*}

We provide confusion matrices for +F+P T5-large models trained on Split 3 with different seeds in Figures \ref{fig:action_conf_matrix_large_d31-1_seed0_split3}, \ref{fig:action_conf_matrix_large_d31-1_seed1_split3}, and \ref{fig:action_conf_matrix_large_d31-1_seed2_split3}. The equivalent figures for T5-base models are \ref{fig:action_conf_matrix_d31-1_seed0_split3}, \ref{fig:action_conf_matrix_d31-1_seed1_split3}, and  \ref{fig:action_conf_matrix_d31-1_seed2_split3}.

\section{Flow prediction confusion matrices}
\label{sec:flow-prediction-confusion-matrices}

\begin{figure*}[hbt!]
    \centering
    \includegraphics[width=\textwidth]{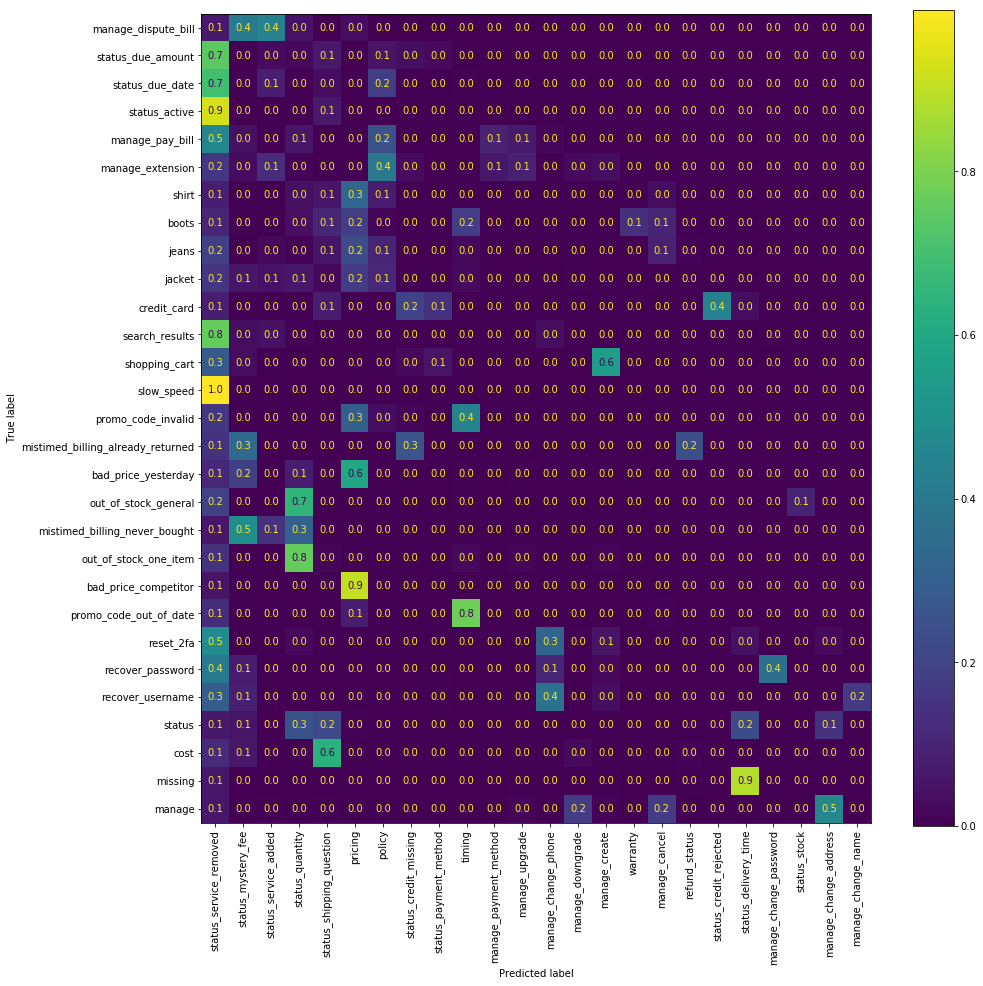}
    \caption{Flow confusion matrix for T5-large finetuned +$\emptyset$ model, seed:42, Split 3. Only rows and columns with significantly non-zero values have been preserved.}
    \label{fig:flow_conf_matrix_large_d16-1_seed2_split3}
\end{figure*}

\begin{figure*}[hbt!]
    \centering
    \includegraphics[width=\textwidth]{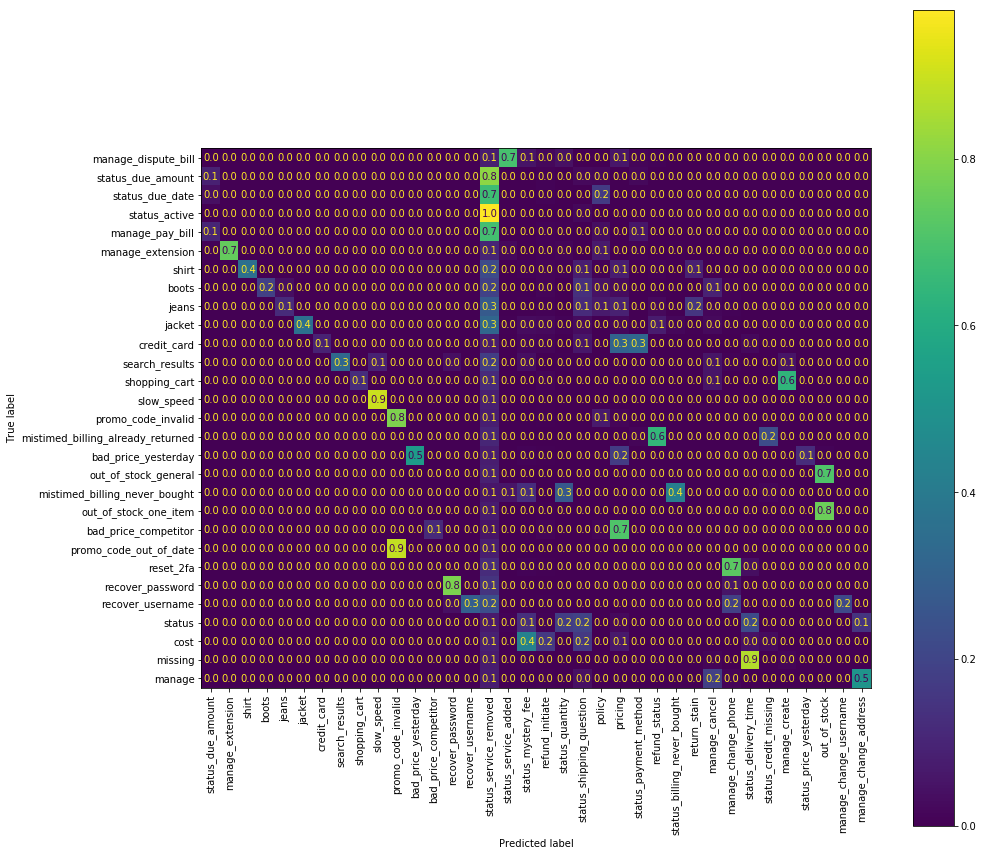}
    \caption{Flow confusion matrix for T5-large finetuned +L model, seed:42, Split 3. Only rows and columns with significantly non-zero values have been preserved.}
    \label{fig:flow_conf_matrix_large_d16-1a_seed2_split3}
\end{figure*}

\begin{figure*}
    \centering
    \includegraphics[width=0.98\textwidth]{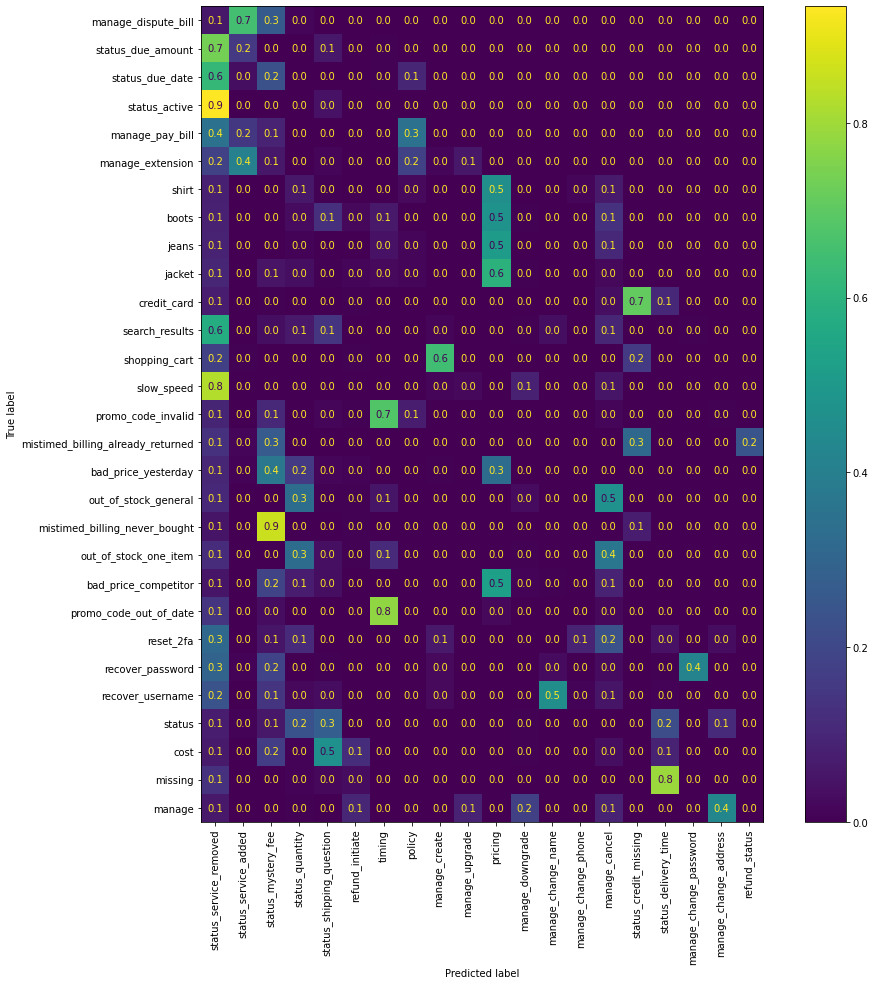}
    \caption{Flow confusion matrix for T5-base finetuned +$\emptyset$ model, seed:42, Split 3. Only rows and columns with significantly non-zero values have been preserved.}
    \label{fig:flow_conf_matrix_d16-1_seed2_split3}

\end{figure*}

\begin{figure*}[hbt!]
    \centering
    \includegraphics[width=\textwidth]{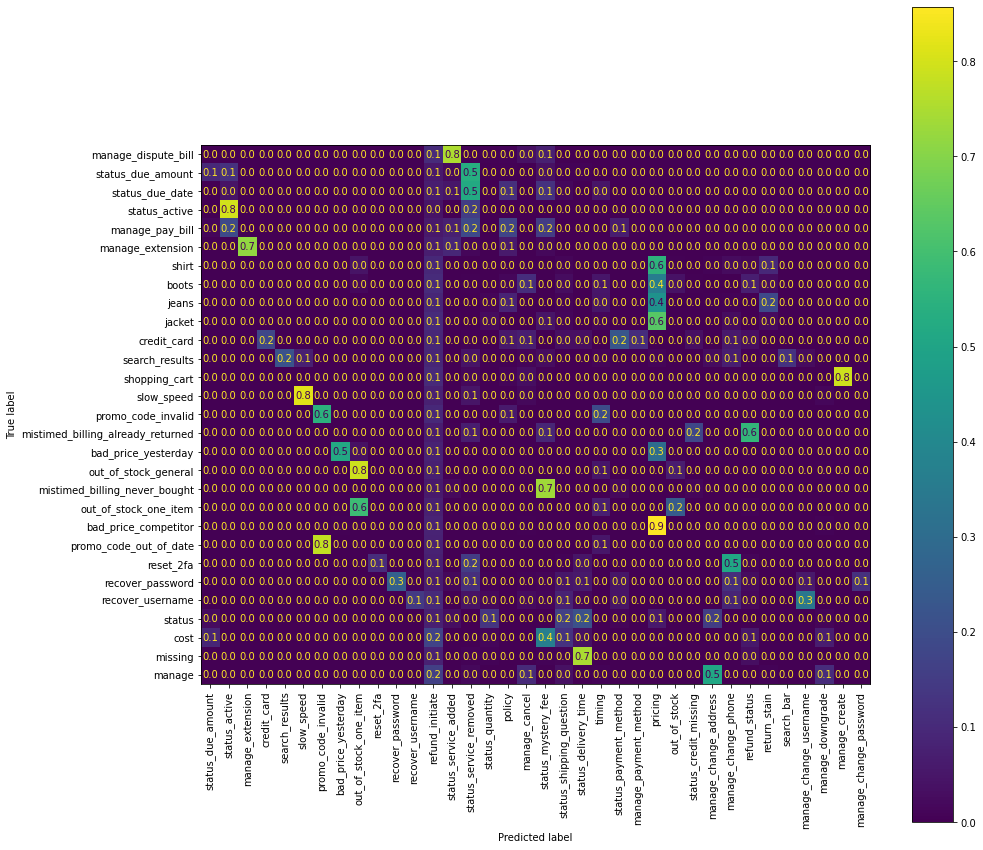}
    \caption{Flow confusion matrix for T5-base finetuned +L model, seed:42, Split 3. Only rows and columns with significantly non-zero values have been preserved.}
    \label{fig:flow_conf_matrix_d16-1a_seed2_split3}
\end{figure*}

We provide flow prediction confusion matrices for +$\emptyset$ and +L in Figures \ref{fig:flow_conf_matrix_large_d16-1_seed2_split3} and \ref{fig:flow_conf_matrix_large_d16-1a_seed2_split3}.
The equivalent figures for T5-base models are \ref{fig:flow_conf_matrix_d16-1_seed2_split3} and  \ref{fig:flow_conf_matrix_d16-1a_seed2_split3}.

\end{document}